\documentclass[10pt,twocolumn,letterpaper]{article}

\usepackage{cvpr}              

\usepackage{graphicx}
\usepackage{amsmath}
\usepackage{amssymb}
\usepackage{amsthm}
\usepackage{booktabs}
\usepackage{multirow}
\usepackage{subcaption}
\usepackage[table,colortbl]{xcolor}
\usepackage{dsfont}
\usepackage{bbm}
\usepackage[makeroom]{cancel}
\usepackage{mathtools}
\usepackage{stfloats}

\usepackage[makeroom]{cancel}
\usepackage{arydshln}
\usepackage[accsupp]{axessibility}

\usepackage[pagebackref,breaklinks,colorlinks]{hyperref}

\usepackage[capitalize]{cleveref}
\crefname{section}{Sec.}{Secs.}
\Crefname{section}{Section}{Sections}
\Crefname{table}{Table}{Tables}
\crefname{table}{Tab.}{Tabs.}

\def\cvprPaperID{6566} 
\def\confName{CVPR}
\def\confYear{2022}

\begin{document}

\title{PCA-Based Knowledge Distillation Towards Lightweight \\and Content-Style Balanced Photorealistic Style Transfer Models}

\author{Tai-Yin Chiu\\
University of Texas at Austin\\
{\tt\small chiu.taiyin@utexas.edu}
\and
Danna Gurari\\
University of Colorado Boulder\\
{\tt\small danna.gurari@colorado.edu }
}

\twocolumn[{%
\renewcommand\twocolumn[1][]{#1}%
\maketitle
\vspace{-1.5em}
\begin{center}
    \centering
    \captionsetup{type=figure}
    \includegraphics[width=\textwidth]{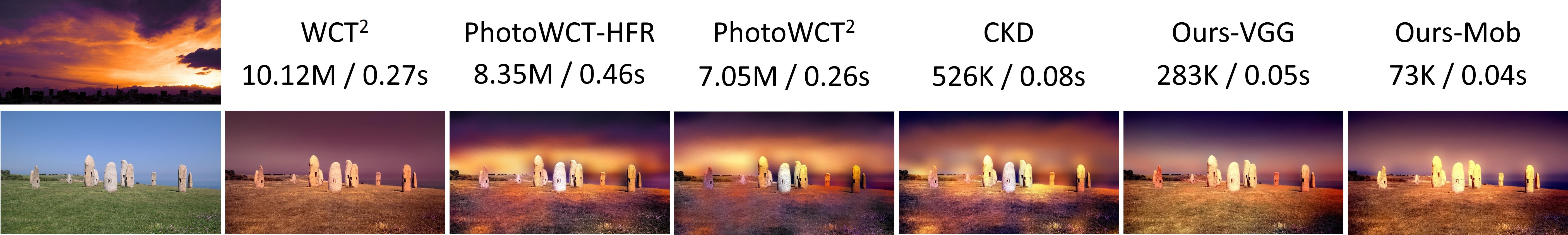}
    \vspace{-1.5em}
    \captionof{figure}{Exemplification of our PCA-based knowledge distillation resulting in models that are more lightweight, faster, and achieve a better content-style balance for photorealistic style transfer than existing models. We demonstrate its versatility by reporting results for when our approach is applied to two backbone architectures: VGG (Ours-VGG) and MobileNet (Ours-Mob). Compared to WCT$^2$, our smallest model uses only $0.7\%$ of the parameters (73K vs. 10.12M) while producing results over 6x faster (0.04s vs. 0.27s).  Additionally, our distilled models consistently transfer stronger style effects than WCT$^2$~\cite{yoo2019photorealistic} and preserve content better than PhotoWCT-HFR (an improved variant of PhotoWCT~\cite{li2018closed}), PhotoWCT$^2$~\cite{chiu2021photowct2}, and a CKD~\cite{wang2020collaborative} distilled model. }
    \label{fig:banner}
\end{center}%
}]

\begin{abstract}
    Photorealistic style transfer entails transferring the style of a reference image to another image so the result seems like a plausible photo. Our work is inspired by the observation that existing models are slow due to their large sizes. We introduce PCA-based knowledge distillation to distill lightweight models and show it is motivated by theory. To our knowledge, this is the first knowledge distillation method for photorealistic style transfer. Our experiments demonstrate its versatility for use with different backbone architectures, VGG and MobileNet, across six image resolutions. Compared to existing models, our top-performing model runs at speeds 5-20x faster using at most 1\% of the parameters.  Additionally, our distilled models achieve a better balance between stylization strength and content preservation than existing  models. To support reproducing our method and models, we share the code at \textit{https://github.com/chiutaiyin/PCA-Knowledge-Distillation}. 
\end{abstract} 
\vspace{-1em}

\section{Introduction}
\label{sec:introduction}

Photorealistic style transfer is the task of rendering an image (content image) in the style of a reference image (style image) to create a photorealistic result. Examples are shown in \cref{fig:banner}. A key challenge the community has focused on tackling since the seminal neural network-based algorithm for this task~\cite{luan2017deep} has been how to simultaneously achieve a good balance between stylization strength and content preservation while running fast to better support practical applications~\cite{li2018closed,yoo2019photorealistic,chiu2021photowct2,xia2020joint}. 

\begin{figure}[t]
    \centering
    \includegraphics[width=\linewidth]{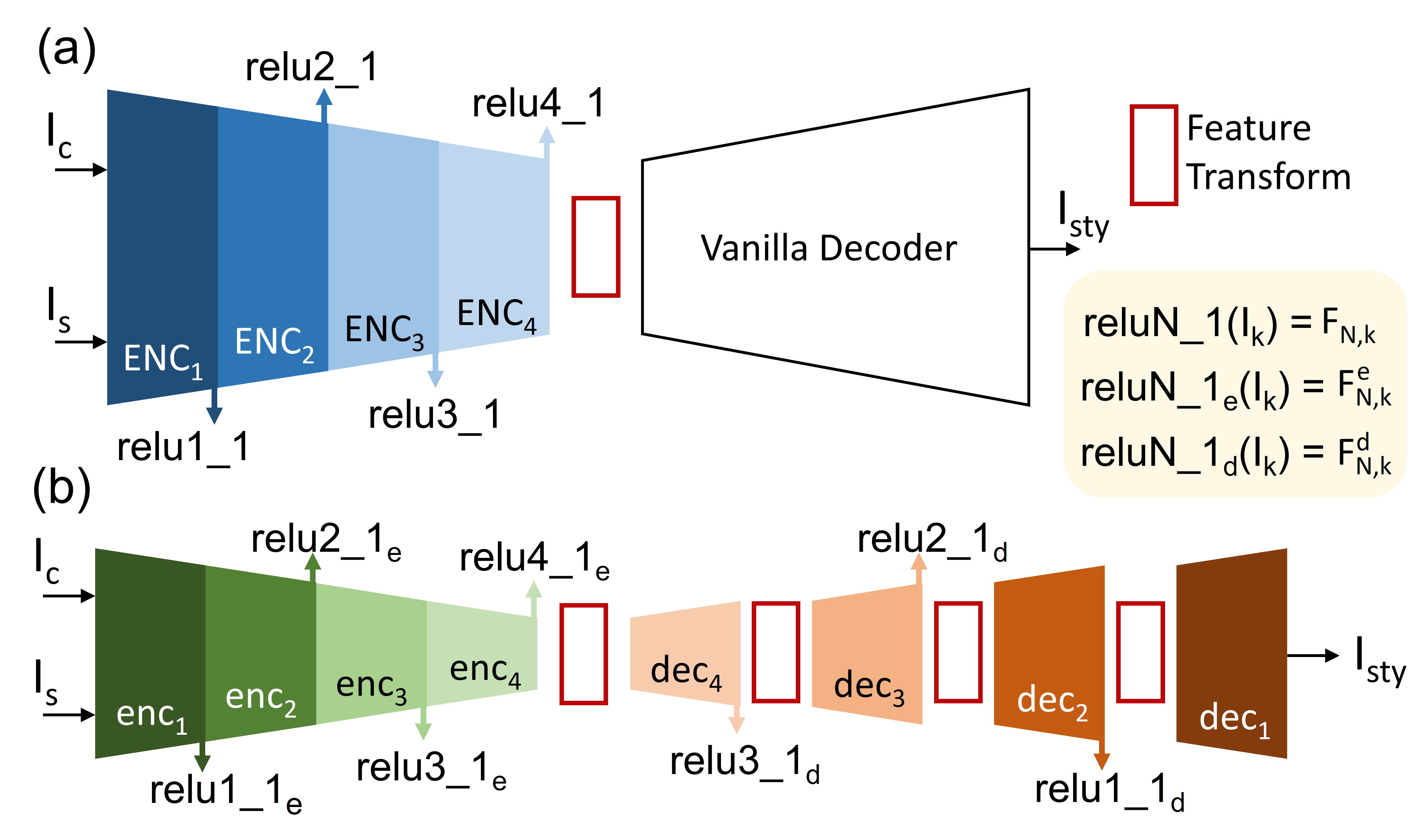}
    \vspace{-1.5em}
    \caption{(a) The basic autoencoder framework for photorealistic style transfer. It uses the encoder $\textit{ENC}$ (VGG-19 from the input to $\textit{relu4\_1}$ layer here) to extract features from the content and style images $I_c$ and $I_s$, applies a feature transformation (usually ZCA feature transformation~\cite{li2017universal}) to adapt the content feature $\mathbf{F}_{4,c}$ with respect to the style feature $\mathbf{F}_{4,s}$, and decodes the adapted feature to a stylized image $I_{sty}$. (b) The $\textit{enc}$-$\textit{dec}$ autoencoder resulting from our method. $\textit{enc}$ is distilled from $\textit{ENC}$ and follows its structure but uses a smaller channel length at each layer. The $\textit{reluN\_1}_e$ layer in $\textit{enc}$ is the counterpart layer of $\textit{reluN\_1}$. If without feature transformations, the feature $\mathbf{F}_{N,k}^d$ of an input image $I_k$ from the $\textit{reluN\_1}_d$ layer in $\textit{dec}$ reproduces the feature $\mathbf{F}_{N,k}^e$ generated at $\textit{reluN\_1}_e$. When performing stylization with feature transformations, our model sequentially adapts the coarse $\textit{relu4\_1}_e$ content feature to the fine $\textit{relu1\_1}_d$ content feature. }
    \label{fig:frameworks}
\end{figure}

The status quo for modern photorealistic style transfer models is to use autoencoders. As illustrated in \cref{fig:frameworks}(a), the basic model uses a pre-trained VGG-19~\cite{simonyan2014very}\footnote{VGG-19 is favored in artistic~\cite{gatys2015neural,johnson2016perceptual,chen2016fast,huang2017arbitrary,ghiasi2017exploring,li2019learning,li2017universal,sheng2018avatar} and photorealistic~\cite{luan2017deep,li2018closed,yoo2019photorealistic,an2020ultrafast,chiu2021photowct2,xia2020joint} style transfer research due to its simple architecture with no complex multiple branches and residual modules, making VGG features easier to interpret for content and style.} as the encoder to extract content and style features, then a feature transformation to adapt the content feature with respect to the style feature, and finally a decoder to invert the adapted feature to a stylized image.  A limitation of this framework is that the speed of autoencoder-based approaches is limited by the large size of the 
VGG-19 backbone.

This size limitation is further amplified in state-of-the-art models, as they extend the basic autoencoder framework by using multiple feature transformations to better capture style effects.  For instance, PhotoWCT~\cite{li2018closed} and PhotoWCT$^2$~\cite{chiu2021photowct2} perform coarse-to-fine feature transformations to sequentially adapt the coarse content feature (i.e., $\textit{relu4\_1}$ content feature from VGG) to the fine content feature (i.e., $\textit{relu1\_1}$ content feature from VGG) with respect to the corresponding style features. This means the fine style feature is added on top of the coarse style feature and so can produce strong style effects. However, as shown in \cref{fig:banner}, the strong style effects may in turn introduce artifacts which ruin the content. Other models such as WCT$^2$~\cite{yoo2019photorealistic} and PhotoNAS~\cite{an2020ultrafast} perform fine-to-coarse feature transformations to adapt the fine content feature first and the coarse later to transfer weaker style effects and avoid artifacts. However, as shown in \cref{fig:banner}, the style effects may be poorly captured since adding the coarse style later partially overshadows the former fine style.  Altogether, we observe that the state-of-the-art photorealistic style transfer models not only suffer from large sizes and so slow speeds, but also poor balance between content preservation and stylization strength.

To address these issues, (1) we propose a \emph{PCA-based knowledge distillation}, which we motivate from PCA theory~\cite{PCA}, to distill the most important knowledge for style representation from a source model to a \emph{smaller encoder}. To demonstrate the generalizability of our distillation method, we apply it to two backbone architectures as the source models: VGG-19 and MobileNet~\cite{howard2017mobilenets}. (2) We integrate the blockwise decoder training of PhotoWCT$^2$~\cite{chiu2021photowct2} into our PCA-based knowledge distillation to concurrently implement a \emph{smaller pairing decoder} which reproduces the encoder features to perform coarse-to-fine feature transformations. The resulting model is shown in \cref{fig:frameworks}(b).  Experiments demonstrate that our smaller models reflect better style than WCT$^2$ and PhotoNAS due to coarse-to-fine feature transformations, preserve better content than PhotoWCT and PhotoWCT$^2$ due to knowledge distillation, and incur faster speeds.  These benefits are exemplified in \cref{fig:banner}. To the best of our knowledge, our method is the first knowledge distillation algorithm for photorealistic style transfer.

\section{Related Works}
\label{sec:related_works}

\paragraph{Modern photorealistic style transfer models.}
To overcome the slow speed of the pioneering neural network-based algorithm~\cite{luan2017deep}, most modern photorealistic style transfer models are based on autoencoders~\cite{li2018closed,yoo2019photorealistic,chiu2020iterative,chiu2021photowct2}.\footnote{An exception is a 2020 method~\cite{xia2020joint} which learns to compute a pixelwise linear mapping to map the content pixel values to the stylized pixel values. However, its code is not available.} Yet, as explained in \cref{sec:introduction}, these models are still relatively slow due to their large sizes.  Additionally, such models result in a poor balance between stylization strength and content preservation.  For instance, PhotoWCT~\cite{li2018closed} realizes strong stylization strength by coarse-to-fine feature transformations, but suffers poor content preservation for two reasons: artifacts from strong stylization strength and missing high-frequency details due to its lossy structure.  WCT$^2$~\cite{yoo2019photorealistic} improves the content preservation from PhotoWCT by performing fine-to-coarse feature transformations to transfer weaker style effects in order to reduce artifacts and also introduces wavelet-based skip connections to reinforce the high-frequency component construction. However, sacrificing stylization strength means WCT$^2$ weakly captures the style.  PhotoWCT$^2$~\cite{chiu2021photowct2} combines the merits of PhotoWCT and WCT$^2$: it introduces high-frequency residuals to improve the high-frequency detail construction from PhotoWCT and realizes coarse-to-fine feature transformations to maintain strong stylization strength.  Yet, PhotoWCT$^2$ still can generate artifacts, likely because of the coarse-to-fine feature transformations.  Our experiments demonstrate that models distilled with our PCA knowledge distillation simultaneously address the aforementioned limitations of existing models: our models are smaller, faster, and achieve a better content-style balance.

\vspace{-3mm}
\paragraph{Knowledge distillation for vision tasks.}
Given a neural network (teacher/source model) for a specific task, knowledge distillation (KD) aims to efficiently train a smaller network (student/target model) for the same task by leveraging the knowledge acquired in the source network.

Most KD algorithms address high-level vision tasks. For example, following the pioneering KD algorithm~\cite{hinton2015distilling}, most follow-up algorithms address image classification~\cite{park2019relational,cho2019efficacy,phuong2019towards,tung2019similarity,yim2017gift,mirzadeh2020improved,peng2019correlation,liu2019knowledge,wang2018kdgan,mishra2017apprentice,yuan2020revisiting}.  Others address object detection~\cite{chen2017learning}, semantic segmentation~\cite{liu2019structured} and clothing matching~\cite{liu2020structured}. Yet, these algorithms are not applicable to the low-level photorealistic style transfer task. For example, it is sufficient for prior work that the coarse features of two dog images from a late layer of a distilled image classifier to be similar such that the final classification layer can classify them to the same class (i.e. dog class).  In contrast, photorealistic style transfer should preserve image content and so it requires that the features differ so that the different content in the dog images can be reconstructed.

The closest work to ours is collaborative knowledge distillation (CKD)~\cite{wang2020collaborative} for artistic style transfer.\footnote{There are three prior works indirectly related to ours: (1) optical flow distillation for video artistic style transfer~\cite{chen2020optical}, (2) knowledge transfer via PCA and graph neural networks~\cite{lee2021interpretable}, and (3) PCA-based distilled dense neural network~\cite{seddik2020lightweight}. In (1), only the optical flow information is distilled to the target model, not the style information. The target model learns the style knowledge from a given style image and so is only able to transfer that particular style. In (2), PCA is not used for knowledge distillation but rather visualization and reduction of computational complexity in graph neural networks. In (3), the source and the target models can only be networks consisting of dense layers and so not convolutional layers. Unlike our method, which treats a feature of an image as a collection of vectors and applies PCA to the vectors, (3) needs to acquire the dense features of all images in a dataset and then applies PCA to them. Consequently, (3) is limited to small network structures and small datasets.} However, CKD is different from ours in several respects. First, CKD is empirical without theoretical explanation. Second, with no theory behind, there is no guideline for target model size selection. Third, CKD is poor at preserving content and so is not applicable to photorealistic style transfer.  We experimentally demonstrate the advantages of our PCA-based approach over CKD.

\section{Method}

We now describe our PCA-based algorithm for distilling lightweight, fast, high quality models for photorealistic style transfer. For simplicity, we describe it when used with the most popular backbone architecture for this task: VGG-19. However, to show the generalizability of our approach, we will also apply the algorithm to distill a model from a MobileNet-based source model in the experiments.  


\subsection{Background: how to represent style}
\label{sec:reformulated_nst}
It has been shown that a style representation is valid (i.e., can catch the style of an image) if the distance between style representations of the style image and the stylized image is a Maximum Mean Discrepancy~\cite{li2017demystifying}. Under this scheme, the covariance of an image feature $\mathbf{F}$ extracted at an intermediate layer of VGG-19 is one feasible style representation. Formally, by reshaping $\mathbf{F}$ $\in$ $C\times H\times W$ ($C$, $H$, $W$: channel length, height, and width of $\mathbf{F}$) to a 2D matrix of shape $C\times HW$, the covariance is calculated as $\frac{1}{HW}\mathbf{\bar{F}}\mathbf{\bar{F}}^\mathrm{T}$, where $\mathbf{\bar{F}}$ is the centralized feature $\mathbf{F}-\mu(\mathbf{F})$, and $\mu(\mathbf{F})$ is the mean of $HW$ column vectors of $\mathbf{F}$.  We leverage this style representation in our method.  For brevity, in what follows, we exclude the coefficient of the covariance matrix.

Often multiple layers of a network (e.g. $\textit{reluN\_1}$, $N$ $=$ $1,2,3,4$) are simultaneously used in order to extract a style representation that captures both the coarse and fine style of an image~\cite{gatys2015neural,li2017demystifying}. We follow this approach in our work.


\subsection{PCA-based knowledge distillation}
\label{sec:pca_distillation}
Our PCA-based knowledge distillation algorithm consists of two parts, which we describe below: \emph{global eigenbasis derivation} and \emph{blockwise PCA knowledge distillation}. 

\vspace{-3mm}
\paragraph{Teacher/Source model.} 
We consider VGG-19 from its input to the $\textit{relu4\_1}$ layer as the source model which we name \textit{ENC}. From \textit{ENC} we distill the style knowledge to a smaller target model which we name \textit{enc}, whose structure follows the source model but uses a smaller channel length at each layer. More specifically, the style knowledge will be distilled from the $\textit{reluN\_1}$ ($N$ $=$ $1,2,3,4$) layers of \textit{ENC} to the counterpart layers in \textit{enc} which we call $\textit{reluN\_1}_e$ layers. 

In the following, we let $\mathbf{F}_{N,k}$ $\in$ $C_N\times H_{N,k}W_{N,k}$ be the feature of the image $I_k$ extracted at the $\textit{reluN\_1}$ layer of \textit{ENC} and $\mathbf{F}_{N,k}^e$ $\in$ $C_N^e\times H_{N,k}W_{N,k}$ ($C_N^e$ $\ll$ $C_N$) be the feature extracted at the counterpart $\textit{reluN\_1}_e$ layer of \textit{enc}. The above notations are summarized in \cref{fig:frameworks}. 

\vspace{-3mm}
\paragraph{PCA for style knowledge distillation: image-dependent eigenbases.} 
We represent the style of a photograph $I_k$ using the covariance matrices of the features from the source/target model.  Formally, in the distillation from $\textit{reluN\_1}$ to $\textit{reluN\_1}_e$, we want the covariance of the distilled feature $\mathbf{\bar{F}}_{N,k}^e(\mathbf{\bar{F}}_{N,k}^e)^\mathrm{T}$ to capture the most important information in the covariance of the source feature $\mathbf{\bar{F}}_{N,k}\mathbf{\bar{F}}_{N,k}^\mathrm{T}$ needed for stylization.  Moreover, we want to do this distillation for multiple layers: distill $\textit{reluN\_1}$ layer to the $\textit{reluN\_1}_e$ layer for $N$ $=$ $1,2,3,4$.

This kind of problem is classical dimension reduction with PCA. In PCA, we treat $\mathbf{\bar{F}}_{N,k}$ as a collection of $H_{N,k}W_{N,k}$ numbers of $C_N$-dimensional data points. PCA seeks an orthonormal transformation $\mathbf{W}_{N,k}$ $\in$ $\mathbb{R}^{C_{N}^e\times C_{N}}$ to map these data points to a $C_N^e$-dimensional space, so that the covariance $\mathbf{W}_{N,k}\mathbf{\bar{F}}_{N,k}\mathbf{\bar{F}}_{N,k}^\mathrm{T}\mathbf{W}_{N,k}^\mathrm{T}$ of the mapped feature $\mathbf{W}_{N,k}\mathbf{\bar{F}}_{N,k}$ preserves the essential information in the original covariance $\mathbf{\bar{F}}_{N,k}\mathbf{\bar{F}}_{N,k}^\mathrm{T}$. Mathematically, PCA solves the following optimization problem~\cite{PCA,L1PCA}.
\begin{equation}
    \max_{\mathbf{W}_{N,k}\mathbf{W}_{N,k}^\mathrm{T}=\mathbbm{1}} \mathrm{tr}(\mathbf{W}_{N,k}\mathbf{\bar{F}}_{N,k}\mathbf{\bar{F}}_{N,k}^\mathrm{T}\mathbf{W}_{N,k}^\mathrm{T}),
    \label{eq:pca_local}
\end{equation}
where the constraint $\mathbf{W}_{N,k}\mathbf{W}_{N,k}^\mathrm{T}=\mathbbm{1}$ is the orthonormality property of $\mathbf{W}_{N,k}$. 

Intuitively, we can think of the trace function as a metric to evaluate the information in a covariance matrix and we search an orthonormal basis to maximize the metric value. According to PCA, the $C_N^e$ row vectors of the solution $\mathbf{W}_{N,k}$ to \cref{eq:pca_local} are the $C_N^e$ leading eigenvectors of the covariance $\mathbf{\bar{F}}_{N,k}\mathbf{\bar{F}}_{N,k}^\mathrm{T}$. With $\mathbf{W}_{N,k}$, we train the target model $\textit{enc}$ to produce at the $\textit{reluN\_1}_e$ layer a feature $\mathbf{F}_{N,k}^e$ whose centralized form $\mathbf{\bar{F}}_{N,k}^e$ is $\mathbf{W}_{N,k}\mathbf{\bar{F}}_{N,k}$. 

However, the image-dependent transformations $\mathbf{W}_{N,k}$'s for distillation are problematic, which can be mathematically explained as follows. First, the $\textit{reluN\_1}_e$ layer of the distilled model $\textit{enc}$ defines a $C_N^e$-dimensional space $\mathcal{S}_{C_N^e}$:
\begin{equation}
    \mathcal{S}_{C_N^e} = \{f\in\mathbb{R}^{C_N^e} | \text{ $f$ is a column vector of }\mathbf{\bar{F}}_{N,k}^e, \forall I_k\}.
\end{equation}
Therefore, for any two images $I_{k_1}$ and $I_{k_2}$, the column vectors of $\mathbf{\bar{F}}_{N,k_1}^e$ and $\mathbf{\bar{F}}_{N,k_2}^e$ should be in the same space  $\mathcal{S}_{C_N^e}$. However, the orthonormal transformations $\mathbf{W}_{N,k_1}$ and $\mathbf{W}_{N,k_2}$ calculated from the covariances $\mathbf{\bar{F}}_{N,k_1}\mathbf{\bar{F}}_{N,k_1}^\mathrm{T}$ and $\mathbf{\bar{F}}_{N,k_2}\mathbf{\bar{F}}_{N,k_2}^\mathrm{T}$ may be different, so the column vectors of the mapped feature $\mathbf{W}_{N,k_1}\mathbf{\bar{F}}_{N,k_1}$ may reside in a different $C_N^e$-dimensional space than the column vectors of $\mathbf{W}_{N,k_2}\mathbf{\bar{F}}_{N,k_2}$, which conflicts with the former argument. Thus, if we distill with the \emph{image-dependent eigenbases} $\mathbf{W}_{N,k}$'s, it results in a \emph{sub-optimal space that does not catch the style of any images well}. We will show this in \cref{sec:content_style_balance}.

\vspace{-3mm}
\paragraph{PCA for style knowledge distillation: global eigenbases (i.e., image-independent).} 
We instead introduce a global, image-independent eigenbasis $\mathbf{W}_{N,g}$ $\in$ $\mathbb{R}^{C_N^e\times C_N}$.  This implementation defines a unique $C_N^e$-dimensional space that on average catches the style of different images well. 

Our solution entails modifying \cref{eq:pca_local} towards solving the following optimization problem for $\mathbf{W}_{N,g}$:
\begin{equation}
\small
    \max_{\mathbf{W}_{N,g}\mathbf{W}_{N,g}^\mathrm{T}=\mathbbm{1}} \frac{1}{M} \sum_{k=1}^M \mathrm{tr}(\mathbf{W}_{N,g}\mathbf{\bar{F}}_{N,k}\mathbf{\bar{F}}_{N,k}^\mathrm{T}\mathbf{W}_{N,g}^\mathrm{T}),
    \label{eq:pca_global}
\end{equation}
where $M$ represents the number of photographs in a virtually infinitely large image dataset, and the solution of $\mathbf{W}_{N,g}$ is the eigenbasis of $\frac{1}{M}\sum_{k=1}^M\mathbf{\bar{F}}_{N,k}\mathbf{\bar{F}}_{N,k}^\mathrm{T}$, which is not analytically attainable. While \cref{eq:pca_global} seems approximately solvable using mini-batch gradient descent to minimize the loss $-\sum_{I_k\in\mathcal{B}_t} \mathrm{tr}(\mathbf{W}_{N,g}\mathbf{\bar{F}}_{N,k}\mathbf{\bar{F}}_{N,k}^\mathrm{T}\mathbf{W}_{N,g}^\mathrm{T})/|\mathcal{B}_t|$, where $\mathcal{B}_t$ is a batch of sampled images at the $t$-th iteration of gradient descent, the minus of a trace as a loss function makes the gradient descent process unstable.  This is because such a loss has no lower bound and so a gradient descent algorithm is prone to keep minimizing the loss but ignore the constraint $\mathbf{W}_{N,g}\mathbf{W}_{N,g}^\mathrm{T}=\mathbbm{1}$.

To bypass it, we rewrite \cref{eq:pca_global} as the equivalent form\footnote{Derivation of the equivalence is in the Supplementary Materials.}:
\begin{equation}
    \min_{\mathbf{W}_{N,g}\mathbf{W}_{N,g}^\mathrm{T}=\mathbbm{1}}\frac{1}{M} \sum_{k=1}^M ||\mathbf{W}_{N,g}^\mathrm{T}\mathbf{W}_{N,g}\mathbf{\bar{F}}_{N,k} - \mathbf{\bar{F}}_{N,k}||^2_2,
    \label{eq:pca_global_L2}
\end{equation}
where $\mathbf{W}_{N,g}^\mathrm{T}\mathbf{W}_{N,g}\mathbf{\bar{F}}_{N,k}$ is the reconstructed feature of $\mathbf{\bar{F}}_{N,k}$ from the mapped feature $\mathbf{W}_{N,g}\mathbf{\bar{F}}_{N,k}$. The meaning behind the equivalence is that the orthonormal basis $\mathbf{W}_{N,g}$ that on average \emph{maximizes the covariance} information in the $C_N^e$-dimensional space should also on average \emph{minimize the reconstruction loss} in the original $C_N$-dimensional space. Unlike the trace function in \cref{eq:pca_global} which is unbounded, the L2-norm of \cref{eq:pca_global_L2} has a lower bound of zero and so \cref{eq:pca_global_L2} is approximately solvable with mini-batch gradient descent. Moreover, since $\mathbf{W}_{N,g}$'s ($N$ $=$ $1,2,3,4$ for $\textit{reluN\_1}$ layers) are independent, they can be solved together with the same batch of sampled images at an iteration of gradient descent.

\begin{figure*}[!t]
\centering
     \begin{subfigure}[b]{0.497\textwidth}
         \centering
         \includegraphics[width=\textwidth]{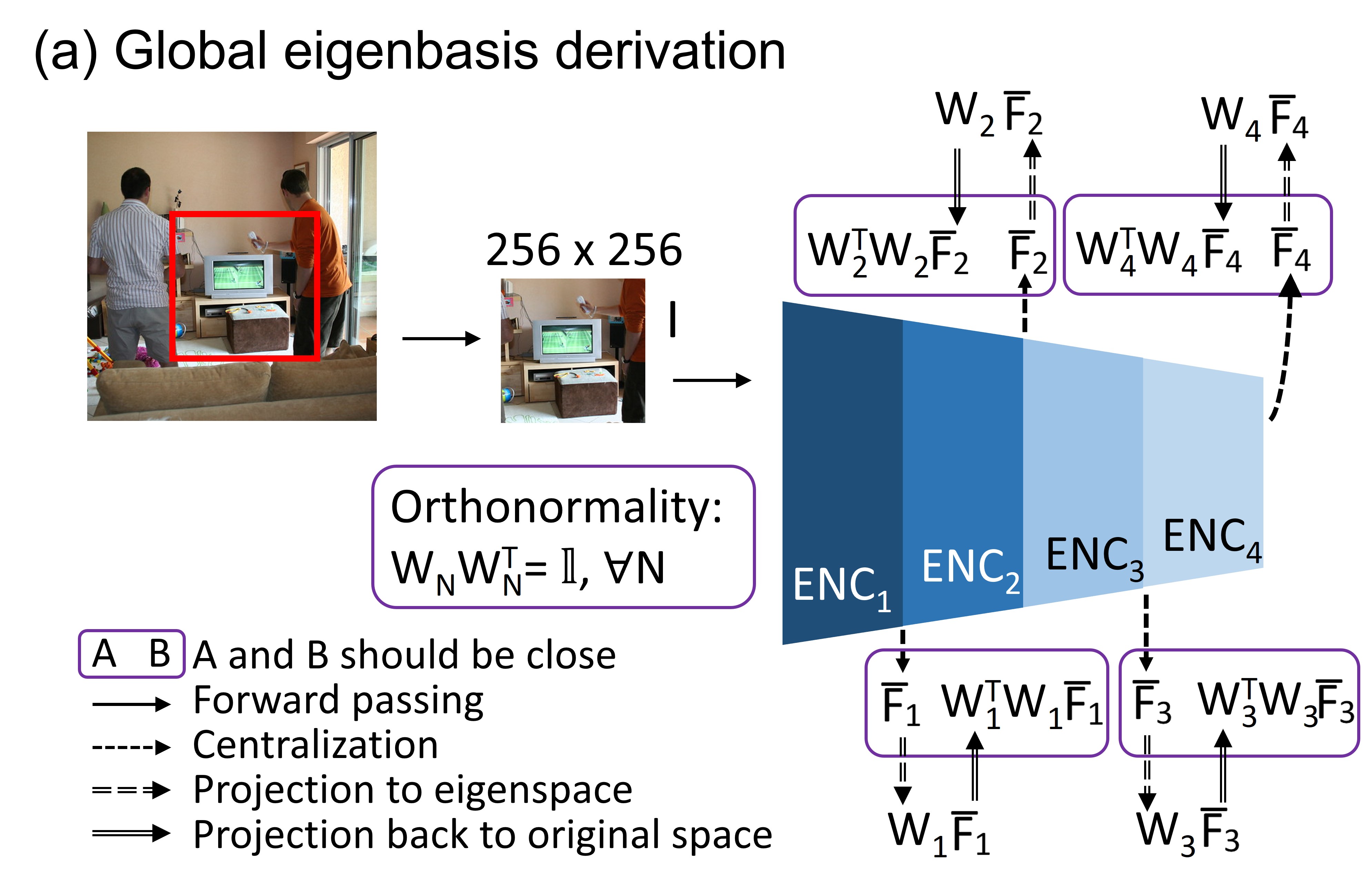}
     \end{subfigure}
     \hfill
     \begin{subfigure}[b]{0.497\textwidth}
         \centering
         \includegraphics[width=\textwidth]{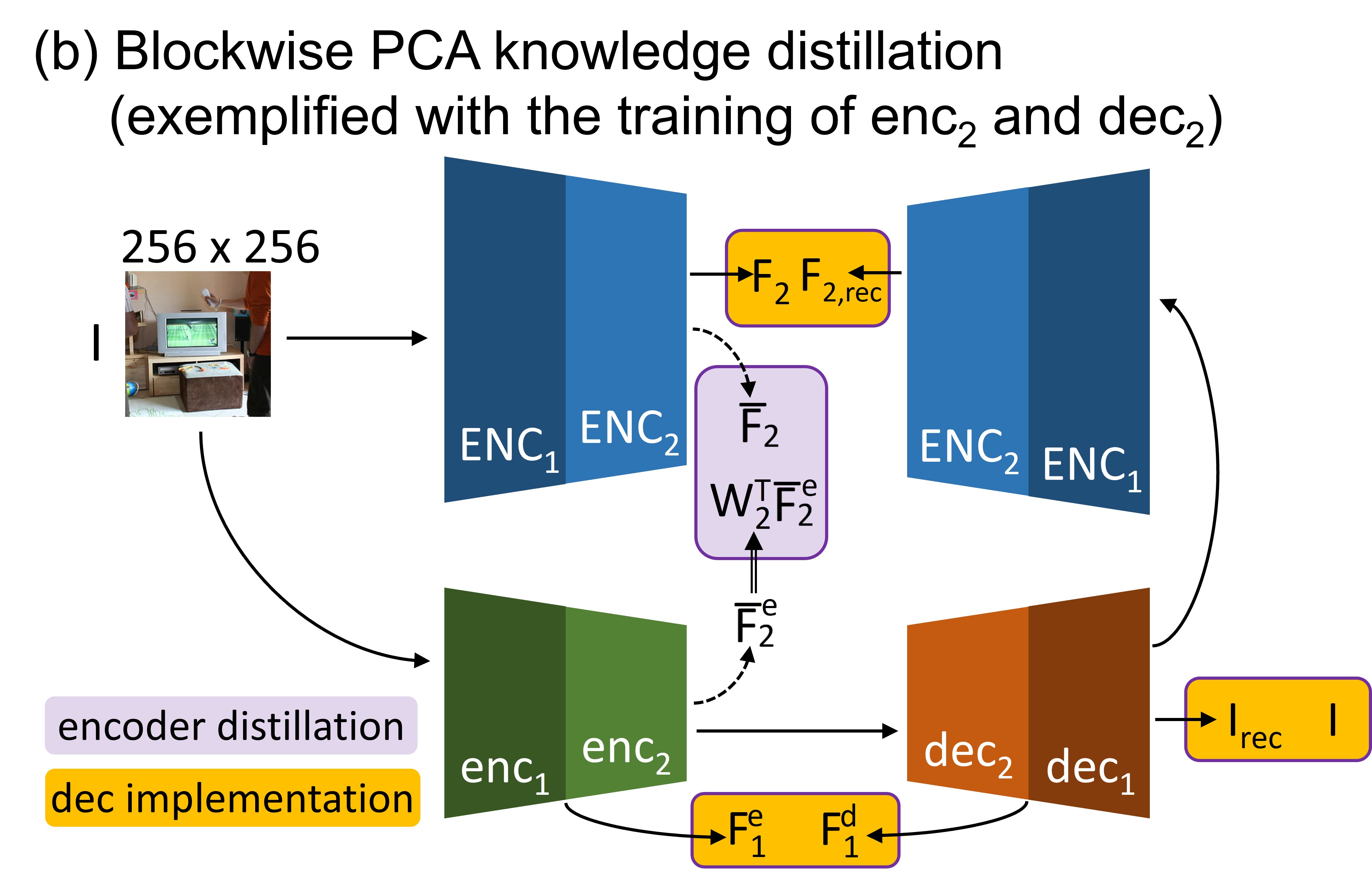}
     \end{subfigure}
     \vspace{-1.5em}
     \caption{PCA-based knowledge distillation for photorealistic style transfer consists of two steps: (a) global eigenbases ($\mathbf{W}_N$, $\mathbf{N}=1,2,3,4$) derivation and (b) blockwise PCA knowledge distillation. As explained for \cref{eq:pca_global_L2}, the $\mathbf{W}_N$ that on average maximizes the style information in the distilled space also minimizes the feature reconstruction loss $||\mathbf{\bar{F}}_N-\mathbf{W}_N^\mathrm{T}\mathbf{W}_N\mathbf{\bar{F}}_N||^2_2$. With $\mathbf{W}_N$'s, style knowledge is blockwisely distilled from the source model block $\textit{ENC}_N$ to the target model block $\textit{enc}_N$ in the order from $N=1$ to $N=4$ by minimizing the encoder distillation loss (\cref{eq:distillation_loss}) and the decoder implementation loss (\cref{eq:decoder_loss}).} 
     \label{fig:knowledge_distillation}
\end{figure*}

To summarize, to derive $\mathbf{W}_{N,g}$'s ($N$ $=$ $1,2,3,4$), we use mini-batch gradient descent to solve \cref{eq:pca_global_L2} for four $\mathbf{W}_{N,g}$'s at once: At the $t$-th iteration of gradient descent, we sample a batch $\mathcal{B}_t$ for the following minimization problem:
\begin{equation}
\small
    \boxed{\min_{\substack{\mathbf{W}_{N,g}\mathbf{W}_{N,g}^\mathrm{T}=\mathbbm{1}\\N \in \{1,2,3,4\}}}\frac{1}{|\mathcal{B}_t|}\sum_{N=1}^4 \sum_{I_k\in\mathcal{B}_t} ||\mathbf{W}_{N,g}^\mathrm{T}\mathbf{W}_{N,g}\mathbf{\bar{F}}_{N,k} - \mathbf{\bar{F}}_{N,k}||^2_2}
    \label{eq:pca_global_L2_batch}
\end{equation}
and calculate the gradient of the objective in \cref{eq:pca_global_L2_batch} to update $\mathbf{W}_{N,g}$'s. In our implementation, we use the MS-COCO dataset with random crop data augmentation, a batch size $|\mathcal{B}_t|$ of 8, and train $\mathbf{W}_{N,g}$'s for five epochs. \cref{fig:knowledge_distillation}(a) exemplifies the derivation with a batch of one image.  

With global eigenbases $\mathbf{W}_{N,g}$'s ($N$ $=$ $1,2,3,4$), we can distill style information from $\textit{reluN\_1}$ layers of the source model $\textit{ENC}$ to $\textit{reluN\_1}_e$ layers of the target model $\textit{enc}$.

\vspace{-3mm}
\paragraph{Blockwise PCA knowledge distillation.}
To realize coarse-to-fine feature transformations in a distilled model, in addition to distilling the style information to the distilled encoder $\textit{enc}$, our approach also must implement a pairing decoder $\textit{dec}$ to reproduce encoder features.  We now describe our distillation approach, which integrates PCA knowledge distillation with the blockwise training strategy for the decoder in PhotoWCT$^2$~\cite{chiu2021photowct2} to achieve this.  

A schematic diagram of our $\textit{enc}$-$\textit{dec}$ model is shown in \cref{fig:frameworks}. We split the encoder $\textit{enc}$ into a series of blocks $\{\textit{enc}_1$, $\textit{enc}_2$, $\textit{enc}_3$, $\textit{enc}_4\}$ where the output of $\textit{enc}_N$ is the $\textit{reluN\_1}_e$ layer, and the decoder $\textit{dec}$ into a series of blocks $\{\textit{dec}_4$, $\textit{dec}_3$, $\textit{dec}_2$, $\textit{dec}_1\}$ where the output of $\textit{dec}_N$ is the $\textit{relu(N-1)\_1}_d$ layer which reproduces the $\textit{relu(N-1)\_1}_e$ feature. That is, the decoder takes as input the $\textit{relu4\_1}_e$ feature from $\textit{enc}$ to progressively reproduce $\textit{relu3\_1}_e$, $\textit{relu2\_1}_e$, and $\textit{relu1\_1}_e$ features and the reconstructed image.  During stylization, ZCA feature transformations~\cite{li2017universal,chiu2019understanding} are placed at the $\textit{relu4\_1}_e$ layer and the $\textit{reluN\_1}_d$ layers ($N$ $=$ $3,2,1$) to sequentially adapt the coarse $\textit{relu4\_1}_e$ content feature to the fine $\textit{relu1\_1}_d$ content feature with respect to the corresponding style features.  

To implement our $\textit{enc}$-$\textit{dec}$ model, each pair of $\textit{enc}_N$ and $\textit{dec}_N$ are trained together with other pairs fixed by minimizing the encoder distillation loss $\mathcal{L}_{enc}^N$ and the decoder implementation loss $\mathcal{L}_{dec}^N$.  The four pairs are trained sequentially in the order from $N$ $=$ $1$ to $N$ $=$ $4$. We exemplify training the pair of $\textit{enc}_2$ and $\textit{dec}_2$ in \cref{fig:knowledge_distillation}(b).

In the encoder distillation, given the image $I_k$ we want to train the encoder block $\textit{enc}_N$ to make its centralized output $\mathbf{\bar{F}}_{N,k}^e$ close to the feature $\mathbf{W}_{N,g}\mathbf{\bar{F}}_{N,k}$, which is mapped from the centralized output $\mathbf{\bar{F}}_{N,k}$ of $\textit{ENC}_N$ by the global eigenbasis $\mathbf{W}_{N,g}$. Instead of taking $||\mathbf{\bar{F}}_{N,k}^e-\mathbf{W}_{N,g}\mathbf{\bar{F}}_{N,k}||^2_2$ as the loss function, we observe the following feature reconstruction loss for encoder distillation converges better:
\begin{equation}
    \mathcal{L}_{enc}^N(I_k) = ||\mathbf{W}_{N,g}^\mathrm{T}\mathbf{\bar{F}}_{N,k}^e - \mathbf{\bar{F}}_{N,k}||^2_2.
    \label{eq:distillation_loss}
\end{equation}
This loss also effectively makes $\textit{enc}_N$ produce $\mathbf{\bar{F}}_{N,k}^e$ equal to $\mathbf{W}_{N,g}\mathbf{\bar{F}}_{N,k}$.\footnote{We prove this in the Supplementary Materials.} We suspect the better convergence is due to its stronger constraint: $\mathcal{L}_{enc}^N(I_k)$ imposes $C_N H_{N,k} W_{N,k}$ scalar constraints, while $||\mathbf{\bar{F}}_{N,k}^e-\mathbf{W}_{N,g}\mathbf{\bar{F}}_{N,k}||^2_2$ imposes $C_N^e H_{N,k} W_{N,k}$ scalar constraints, which are $C_N/C_N^e$ times ($\sim$10x) looser and might be casually fulfilled.

In the decoder implementation, given the image $I_k$ we want to make the output $\mathbf{F}_{N-1,k}^d$ of  $\textit{dec}_N$ to reproduce the input $\mathbf{F}_{N-1,k}^e$ of $\textit{enc}_N$ and the reconstructed image $I_{k_{rec}}$ from $\textit{dec}_1$ close to the input $I_k$. Collectively, we minimize $\mathcal{L}_{dec}^N$ consisting of three terms:
\begin{equation}
\begin{aligned}
    \mathcal{L}_{dec}^N(I_k) = ~&||\mathbf{F}_{N-1,k}^d-\mathbf{F}_{N-1,k}^e||^2_2\\
    +&||I_{k_{rec}}-I_k||^2_2+||\mathbf{F}_{N,k_{rec}}-\mathbf{F}_{N,k}||^2_2,
    \label{eq:decoder_loss}
\end{aligned}
\end{equation}
where the third term is perceptual loss used to boost the image reconstruction. Note that when $N=1$, there is no first term for feature reproduction.

To summarize, when training the pair of $\textit{enc}_N$ and $\textit{dec}_N$, we solve the following optimization problem:
\begin{equation}
    \boxed{\min_{\textit{enc}_N,~\textit{dec}_N} \mathcal{L}_{enc}^N(I_k) + \mathcal{L}_{dec}^N(I_k).}
\end{equation}

\begin{figure*}[t!]
    \centering
    \includegraphics[width=\textwidth]{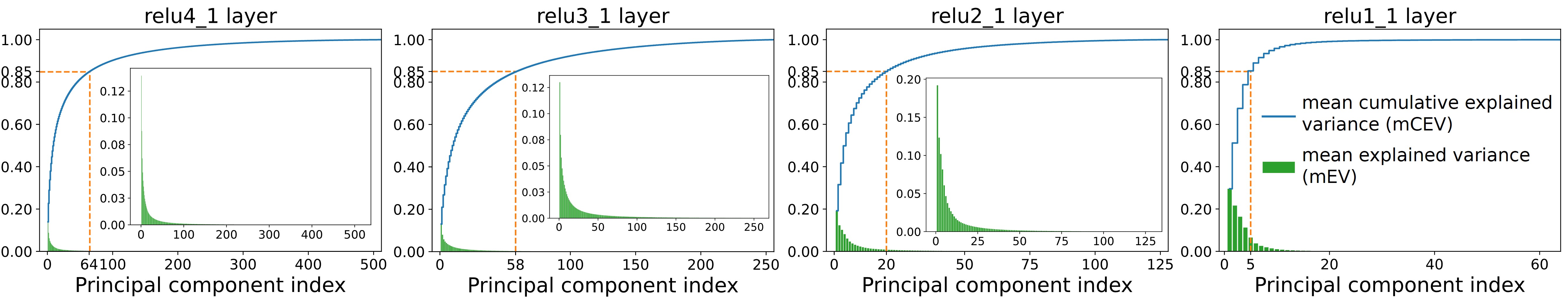}
    \vspace{-1.5em}
    \caption{Mean explained variance (green histogram) and mean cumulative explained variance (blue curve) of the $\textit{reluN\_1}$ features of MS-COCO images. It is observed that on average $85\%$ of the variance of a $\textit{relu4\_1}$, $\textit{relu3\_1}$, $\textit{relu2\_1}$, or $\textit{relu1\_1}$ feature can be explained by $64$, $58$, $20$, or $5$ eigenvectors of the feauture covariance, respectively.}
    \label{fig:explained_variance}
\end{figure*}

\subsection{Channel lengths of the target model}
We now show how we select the channel lengths $C_N^e$'s of the target model $\textit{enc}$. Following the rule of thumb~\cite{artoni2018applying,gajjar2018real,corvucci2015discrimination,odlare2005near,razmkhah2010evaluation} in PCA dimension reduction to keep the most important information, the target layer $\textit{reluN\_1}_e$ of the channel length $C_N^e$ should preserve $85\%$ of the variance information from the source layer $\textit{reluN\_1}$.  The channel length that preserves this amount of information can be determined with a data-driven approach.

For each image $I_k$ in our dataset, we compute the covariance of its feature $\mathbf{F}_{N,k}$ extracted at the $\textit{reluN\_1}$ layer. Let $\sigma_{N,k}^j$ be the $j$-th largest eigenvalue of the covariance associated with the $j$-th principal eigenvector $\mathbf{e}_{N,k}^j$. The $j$-th explained variance (EV) $\sigma_{N,k}^j/\sum_{\alpha=1}^{C_N} \sigma_{N,k}^{\alpha}$ reflects the portion of the feature variance captured by $\mathbf{e}_{N,k}^j$. Then the cumulative EV (CEV) $\sum_{\beta=1}^{C_N'}\sigma_{N,k}^{\beta}/\sum_{\alpha=1}^{C_N} \sigma_{N,k}^{\alpha}$ reflects the feature variance captured by the top $C_N'$ eigenvectors. We use mean cumulative explained variance (mCEV) to determine the value of $C_N^e$. The mean CEV (mCEV) is the average of the CEVs across all images:
\begin{equation}
\small
    \text{mCEV}(C_N') = \frac{1}{M}\sum_{k=1}^M\frac{\sum_{\beta=1}^{C_N'}\sigma_{N,k}^{\beta}}{\sum_{\alpha=1}^{C_N} \sigma_{N,k}^{\alpha}} = \sum_{\beta=1}^{C_N'}\text{mEV}(\beta),
\end{equation}
where $M$ is the number of images in the dataset and mEV($\beta$) $=$ $\frac{1}{M}\sum_{k=1}^M(\sigma_{N,k}^{\beta}/\sum_{\alpha=1}^{C_N} \sigma_{N,k}^{\alpha}$) is the mean $\beta$-th EV. We look for a $C_N^e$ that satisfies $\text{mCEV}(C_N^e)$ $\approx$ $85\%$. We use the MS-COCO~\cite{lin2014microsoft} training set of 118,287 images because of its large number of diversified images.

The mCEV and mEV for each $\textit{reluN\_1}$ layer of VGG-19 are shown in \cref{fig:explained_variance}. We observe that $85\%$ of the variance information in the original 512 (256, 128, 64)-dimensional space of the $\textit{relu4\_1}$ ($\textit{relu3\_1}$, $\textit{relu2\_1}$, $\textit{relu1\_1}$) layer can be explained by, on average, 64 (58, 20, 5) principal components.  In other words, across the four layers, a small percentage of the original number of channels are needed to preserve 85\% of the variance information; i.e., $12.5\%$ ($22.7\%$, $15.6\%$, $7.8\%$). 

Experimentally, we found that setting $C_1^e$ to $5$ with $C_2^e$ set to $20$ hinders the distillation from the $\textit{relu2\_1}$ layer to the $\textit{relu2\_1}_e$ layer. We suspect it is due to the low reduction rate of $7.8\%$ for the $\textit{relu1\_1}_e$ layer. We found experimentally that doubling the reduction rate and setting $C_1^e$ to $10$ overcomes this issue. For our final model, we set the four channel lengths ($C_1^e$, $C_2^e$, $C_3^e$, $C_4^e$) to (10, 20, 58, 64).

\section{Experiments}
We evaluate our models with respect to model size (\cref{sec:model_size}), inference time at different image resolutions (\cref{sec:speed_performance}), and quality of rendered images (\cref{sec:content_style_balance}).  

\subsection{Model size}
\label{sec:model_size}


\paragraph{Our models.}
We apply our PCA knowledge distillation to distill two models from VGG and MobileNet.  We refer to these models as Ours-VGG and Ours-Mob, respectively. 

To assess the benefit of using PCA to derive the channel lengths for a target model, we also evaluate against a PCA-distilled model using the channel lengths empirically selected for CKD~\cite{wang2020collaborative}: $C_1^e=16$, $C_2^e=32$, $C_3^e=64$, and $C_4^e=128$. We call this variant Ours-VGG-CKD.

Following prior work\cite{chiu2021photowct2}, we apply high-frequency residuals (HFR) to all of our models to support good high-frequency detail construction.

\vspace{-3mm}
\paragraph{Baselines.}
We compare to four models.\footnote{We exclude a couple recent models from this experiments section for the following reasons.  Due to the huge size, PhotoNAS~\cite{an2020ultrafast} cannot handle the smallest considered image resolution (HD) in this paper and so is not used for comparison. LST~\cite{li2019learning} is another autoencoder-based method, which is designed for artistic style transfer and is insufficient for photorealistic style transfer, as we show in the Supplementary Materials.}  One model is distilled with the only prior distillation method for style transfer: CKD~\cite{wang2020collaborative}.  We also evaluate three state-of-the-art non-distilled models: WCT$^2$~\cite{yoo2019photorealistic}, PhotoWCT$^2$~\cite{chiu2021photowct2}, and PhotoWCT~\cite{li2018closed}.  To support fair comparison, we apply HFR to reinforce the high-frequency detail construction for those models that lack this feature and so suffer lossy structure for high-frequency details (as discussed in \cref{sec:related_works}): PhotoWCT~\cite{li2018closed} and the CKD model.  We show in the Supplementary Materials that both variants achieve better content preservation than the original versions, with little impact to model size and speed.  

\vspace{-3mm}
\paragraph{Results.}
Results are reported in \cref{tab:model_charateristics}(a). 

Compared to the existing models, our models have far fewer model parameters. For example, Ours-VGG uses only $2.8\%$, $3.4\%$, and $4.0\%$ of the parameters of WCT$^2$, PhotoWCT-HFR, and PhotoWCT$^2$, respectively.  Given that MobileNet is a lightweight alternative to VGG-19,  we observe that Ours-Mob uses fewer parameters than Ours-VGG: it uses only $0.7\%$, $0.9\%$, and $1.0\%$ of the parameters of WCT$^2$, PhotoWCT-HFR, and PhotoWCT$^2$, respectively.

Compared to the baseline distilled model, CKD, Ours-VGG-CKD uses fewer parameters.  We attribute that to the fact that Ours-VGG-CKD uses a single autoencoder while CKD uses a cascade of four autoencoders. 

When comparing the sizes of models distilled using the channel lengths derived with PCA and those empirically selected for CKD~\cite{wang2020collaborative}, Ours-VGG uses $64.3\%$ parameters of Ours-VGG-CKD.  In other words, our theoretically motivated approach results in a more compact model.\footnote{
Note that a certain amount of variance needs to be preserved so that the model compression does not result in an architecture with too little capacity to effectively learn. Through experimentation, we found the lower bound for the variance information percentage is around 75\%, which results in 59K parameters with VGG as the source model. While the distilled model with 75\% of the variance information marginally decreases the content loss from Ours-VGG (2.20e$^6$ vs. 2.25e$^6$) and slightly increases the style loss (4.90e$^4$ vs. 4.77e$^4$), the distillation of this model is unstable and relies on good initialization of parameter values.}

\subsection{Inference time}
\label{sec:speed_performance}
Intuitively, smaller models should enable faster stylization and support larger image resolutions.  We demonstrate these benefits here.  We conduct all experiments on an Nvidia RTX8000 GPU with 48GB memory. For completeness, we report CPU times in the Supplementary Materials.

\vspace{-3mm}
\paragraph{Our models and baselines.}
We test the same models and baselines as used in \cref{sec:model_size}.

\vspace{-3mm}
\paragraph{Dataset.}
We test all models on six resolutions: $1280$$\times$$720$ (HD), $1920$$\times$$1080$ (Full HD), $2560$$\times$$1440$ (Quad HD), $3840$$\times$$2160$ (4K UHD), $5120$$\times$$2880$ (5K), and $7680$$\times$$4320$ (8K). We collect images for testing by downloading an 8K video~\cite{8kvideo} from YouTube, sampling a frame per second to collect 100 frames, and then downsampling each frame to the other lower resolutions.

\setlength{\tabcolsep}{1.8pt}
\begin{table}[t]
    \centering\small
    \def\arraystretch{1.2}
    \begin{tabular}{l c c c c c c c}
    \toprule
        \multirow{2}{*}{Model} & \multirow{2}{*}{(a) Size} & \multicolumn{5}{c}{(b) Inference time} \\
    \cmidrule[0.5pt](lr){3-8}
         & & HD & FHD & QHD & 4K & 5K & 8K \\
    \midrule
    WCT$^2$ & 10.12M & 0.37 & 0.80 & 1.20 & OOM & OOM & OOM\\
    PhotoWCT-HFR & 8.35M & 0.56 & 0.84 & 1.27 & 2.53 & 4.36 & OOM\\
    PhotoWCT$^2$ & 7.05M & 0.32 & 0.45 & 0.69 & 1.26 & 2.14 & OOM \\
    \rowcolor{lightgray!40}[\dimexpr\tabcolsep+0.1pt\relax] CKD & 526K & 0.09 & 0.16 & 0.25 & 0.53 & 0.93 & 2.09\\
    \rowcolor{lightgray!40}[\dimexpr\tabcolsep+0.1pt\relax] Ours-VGG-CKD & 440K & 0.06 & 0.09 & 0.13 & 0.25 & 0.42 & 0.92 \\\noalign{\vskip-0.25pt}
    \rowcolor{lightgray!40}[\dimexpr\tabcolsep+0.1pt\relax] Ours-VGG & \bf{283K} & \bf{0.05} & \bf{0.07} & \bf{0.11} & \bf{0.22} & \bf{0.38} & \bf{0.82} \\\noalign{\vskip-0.25pt}
    \hdashline
    \rowcolor{lightgray!40}[\dimexpr\tabcolsep+0.1pt\relax] Ours-Mob & \bf{73K} & \bf{0.04} & \bf{0.05} & \bf{0.06} & \bf{0.10} & \bf{0.17} & \bf{0.38} \\\noalign{\vskip-0.25pt}
    \bottomrule
    \end{tabular}
    \vspace{-0.5em}
    \caption{Size and inference time of different models, with all distilled models highlighted in light gray. Note that all models except the separated Ours-Mob are distilled from VGG-19. Compared to existing models, our models are smaller and achieve higher speeds.  OOM: Out of memory. Unit: Second/Image.}
    \label{tab:model_charateristics}
\end{table}

\vspace{-3mm}
\paragraph{Results.}
We report the speed of different models for stylizing images of different resolutions in \cref{tab:model_charateristics}. 

Both our PCA-based models, Ours-VGG and Ours-Mob, run the fastest for all considered resolutions of the VGG-based models and of all models, respectively. Compared to the fastest non-distilled model PhotoWCT$^2$, Ours-VGG is consistently 5-6x faster in the resolutions supported by PhotoWCT$^2$, while Ours-Mob is 8x faster in HD image stylization and 12.6x faster in 4K and 5K image stylization.  Compared to the slowest non-distilled model WCT$^2$, Ours-VGG is 7x faster in HD image stylization and 11x faster in FHD and QHD image stylization, while Ours-Mob is 9x, 16x, and 20x faster in HD, FHD, and QHD image stylization, respectively. Compared to the only baseline distilled model, CKD, Ours-VGG is consistently 2-2.5x faster in all resolutions while Ours-Mob can even run 5.3-5.5x faster in 4K, 5K and 8K resolutions.

Regarding supported resolutions, only distilled models can support the highest resolution tested: 8K.  Moreover, WCT$^2$ is not even able to support 5K resolutions. This finding highlights that distilling models to smaller sizes is critical for supporting higher resolution visual data.

\vspace{2mm}
\subsection{Content preservation and stylization strength}
\label{sec:content_style_balance}
We assess the quality of our models' rendered images. 

\vspace{-3mm}
\paragraph{Our models.}
We again evaluate Ours-VGG, Ours-Mob, and Ours-VGG-CKD, which are described in \cref{sec:model_size}.  Additionally, to assess the benefit of using global eigenbases to distill our models, we also evaluate models distilled using image-dependent eigenbases: Ours-VGG-Local and Ours-Mob-Local, where $\textit{Local}$ denotes local eigenbases.

\vspace{-3mm}
\paragraph{Baselines.}
We test the same baselines as used in \cref{sec:model_size}.\footnote{For completeness, we show in the Supplementary Materials that CKD's image quality is not better when using our smaller channel lengths.}  

\vspace{-3mm}
\paragraph{Metrics.}
We adopt the de facto standard metrics to measure content preservation and stylization strength, which were established in ~\cite{gatys2015neural}.  Specifically, given a stylized image $I_t$, we calculate its content loss $||\mathbf{\bar{F}}_{4,c}-\mathbf{\bar{F}}_{4,t}||_2^2$ from the content image $I_c$ and its style loss $\sum_{N=1}^4 ||\text{Cov}(\mathbf{F}_{N,s}) - \text{Cov}(\mathbf{F}_{N,t})||^2_2$ from the style image $I_s$. 

We also assess the quality of an image $I_t$, using standard metrics employed in the image quality assessment community: SSIM~\cite{wang2004image}, FSIM~\cite{zhang2011fsim}, and NIMA~\cite{talebi2018nima}.  While the first two metrics measure the structure similarity between $I_t$ and $I_c$, the third metric evaluates $I_t$ as a standalone image. 

\begin{figure*}[!t]
    \centering
    \includegraphics[width=\textwidth]{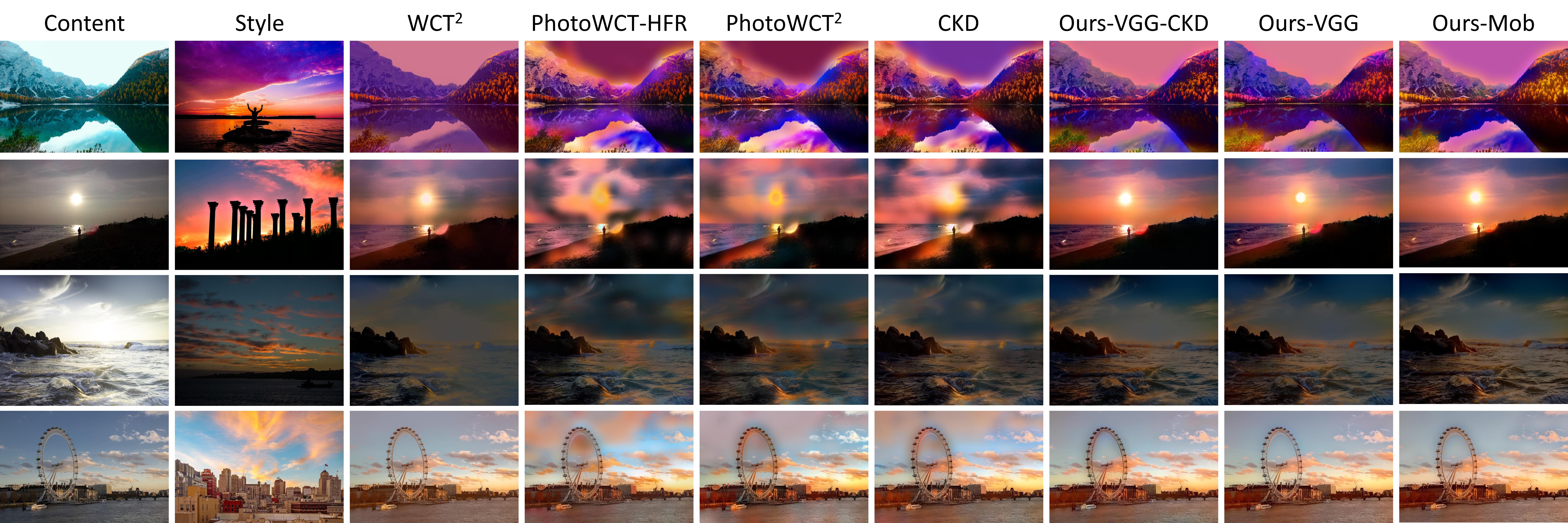}
    \vspace{-1.5em}
    \caption{Our PCA-distilled models result in stronger style effects than WCT$^2$ and better photorealism than the other methods, achieving a better content-style balance. Please see more stylized images in Supplementary Materials.}
    \label{fig:qualitative_results}
    \vspace{-1.0em}
\end{figure*}

\begin{figure}[!t]
    \centering
    \includegraphics[width=\linewidth]{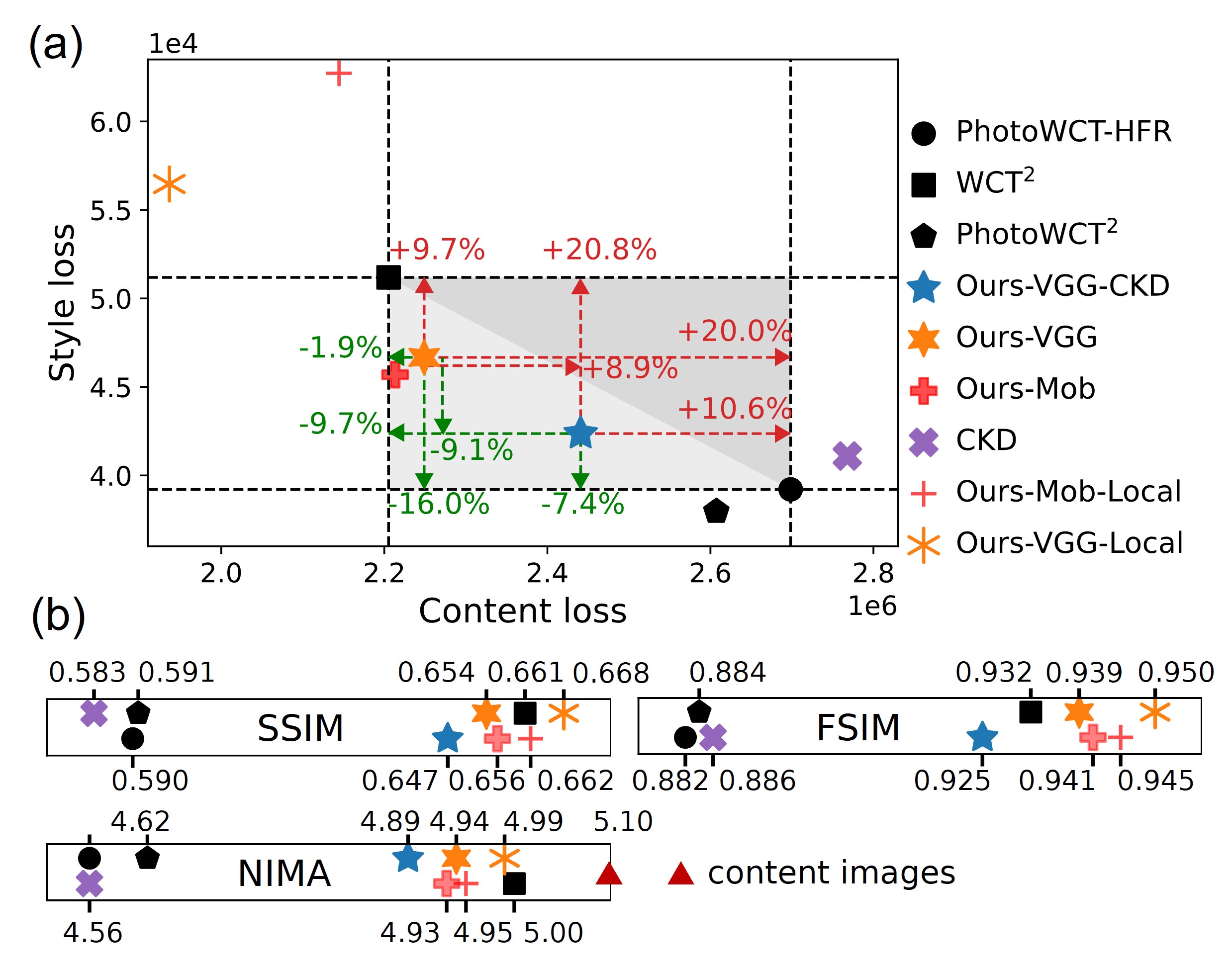}
    \vspace{-1.75em}
    \caption{(a) Content vs. style losses resulting from different models. Our models distilled with global eigenbases (Our-VGG-CKD, Our-VGG, Our-Mob) can trade-off a smaller amount of stylization strength/content from PhotoWCT/WCT$^2$ for a larger gain in content preservation/stylization strength. (b) Quality scores of stylized images resulting from different models. Our models consistently achieve higher scores. }
    \label{fig:content_style_balance}
    \vspace{-1.0em}
\end{figure}

\vspace{-3mm}
\paragraph{Dataset.}
We use PST dataset~\cite{xia2020joint}, the largest publicly available dataset for evaluating photorealistic style transfer models.  It consists of 786 pairs of content and style images, where the resolutions of images are between HD and FHD.


\vspace{-3mm}
\paragraph{Results.}
Qualitative results are shown in \cref{fig:qualitative_results} and quantitative results are shown in \cref{fig:content_style_balance}.

Overall, our models distilled with global eigenbases achieve the best balance of content preservation and stylization strength (\cref{fig:content_style_balance}(a)).  This is evident when examining the two extremes for the content-style trade-off: WCT$^2$ sets the upper bound for style loss since it achieves the weakest stylization strength while PhotoWCT-HFR sets the upper bound for content loss since it achieves the worst content preservation.  We denote the region set by these two extremes in gray.  Lighter gray denotes the region where the trade-off is a smaller amount of stylization strength/content from PhotoWCT-HFR/WCT$^2$ for a larger gain in content preservation/stylization strength.  All our distilled models (Ours-VGG, Ours-Mob, Ours-CKD) achieve a better balance between stylization strength and content preservation, with our most compact model (i.e., Ours-Mob) achieving the best balance.  We suspect this is because PCA filters out marginal pattern information (e.g. cloud contours, ripples) that can cause artifacts while preserving the dominant color information. In contrast, we suspect that the high content loss of CKD reflects that CKD poorly filters out the non-essential pattern information, as shown in \cref{fig:qualitative_results}.

Compared to Ours-VGG-CKD,  Ours-VGG further improves the content preservation by trimming extra style information that results in slight artifacts\footnote{The slight artifacts are not clear in the small resolutions in \cref{fig:qualitative_results}. We show stylized images of 4K+ resolutions in Supplementary Materials where the slight artifacts become apparent.}, which the 9.1\% lower style loss of Ours-VGG-CKD mostly contributes to. This highlights that, with the guidance from the PCA theory, we can properly select the channel lengths for our models to achieve a better content-style balance than when empirically selecting from trials.

The benefit of our PCA distillation strategy is validated by comparing models based on local eigenbases (i.e., Ours-VGG-Local and Ours-Mob-Local) to our models based on global eigenbases.  We observe a stronger stylization strength in our models based on global eigenbases (qualitative comparisons are shown in Supplementary Materials).

Overall, we observe similar trends to those noted above with the quality metrics (\cref{fig:content_style_balance}(b)).  Our models outperform existing distilled and non-distilled models.  

\subsection{Limitation}

While the primary benefit of our work is to demonstrate that we can create ultra-compact models that run very fast, our work also highlight that stylizing images of higher resolutions is now in reach (i.e., 4K resolutions and beyond).  However, we show in the Supplementary Materials that our models occasionally result in artifacts for the larger resolution images, since they contain more high-frequency details which are hotbeds where artifacts form.  Future work should establish a new benchmark dataset for photorealistic stylization that includes larger resolution images. 



\section{Conclusion}
We propose the first knowledge distillation for photorealistic style transfer, which is motivated by PCA theory.  Compared to existing distilled and non-distilled models, our models are smaller, faster, and achieve a better balance between content preservation and stylization strength.


\section*{Supplementary Materials}
This document supplements the main paper with the following.
\begin{enumerate}
    \item Results demonstrating the insufficiency of LST for photorealistic style transfer (supplements Section 2 of the main paper).
    \item Derivation of the equivalence of Equations 3 and 4 in the main paper.
    \item Derivation of the loss objective in Equation 6 in the main paper.
    \item Channel length selection for Ours-Mob model distilled from MobileNet (supplements Section 3.3 in the main paper).
    \item Results demonstrating that high-frequency residuals improve the high-frequency detail construction of PhotoWCT and the original CKD-distilled model (supplements Section 4 of the main paper).
    \item Inference time of different models on the CPU (supplements Section 4.2 of the main paper).
    \item Demonstration that CKD does not improve performance when implemented using our PCA-derived channel lengths instead of the empirical ones (supplements Section 4.3 of the main paper).
    \item Qualitative results to supplement those in Section 4.3 of the main paper.
    \item Results demonstrating that global eigenbases reflect style better than local eigenbases (supplements Section 4.3 of the main paper).
    \item Results of stylized images for 4K+ resolutions from our PCA-distilled models (supplements Section 4.3 of the main paper).
\end{enumerate}

\begin{figure}[t]
    \centering
    \includegraphics[width=\linewidth]{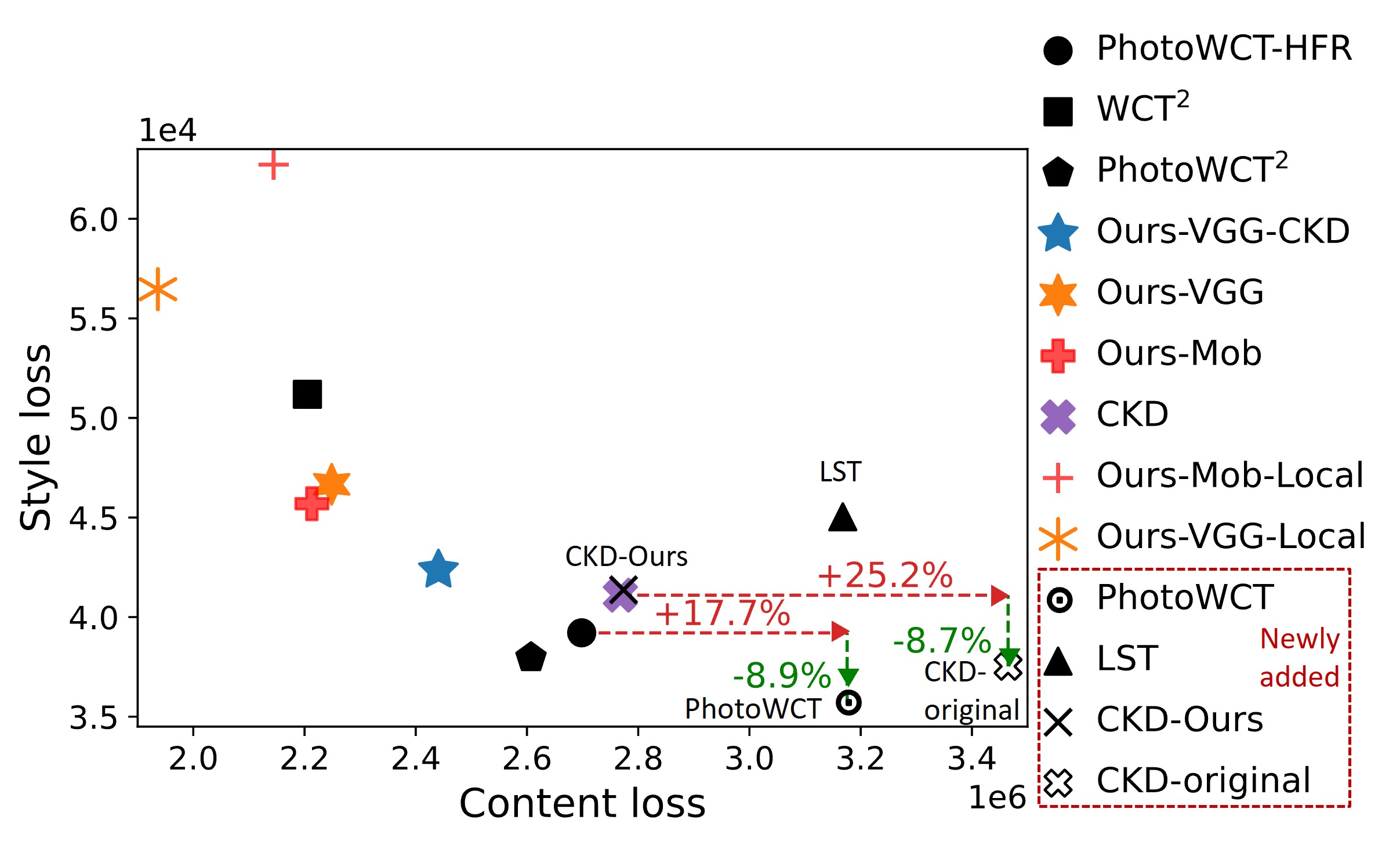}
    \caption{Content and style losses resulting from different models. This figure extends the Figure 5(a) in the main paper. Four newly added points are for LST~\cite{li2019learning}, PhotoWCT~\cite{li2018closed},  CKD-original~\cite{wang2020collaborative}, and CKD-Ours.}
    \label{fig:content_style_losses_all}
\end{figure}

\section*{Insufficiency of LST for photorealistic style transfer}
In the Section 2 of the main paper, we mention LST~\cite{li2019learning}, an autoencoder-based model for artistic style transfer, is claimed to be capable of photorealistic style transfer. However, it does not provide any quantitative analysis for that assertion. Here we show LST is not sufficient for photorealistic style transfer by providing both quantitative and qualitative results.

\begin{figure*}[t]
    \centering
    \includegraphics[width=0.9\textwidth]{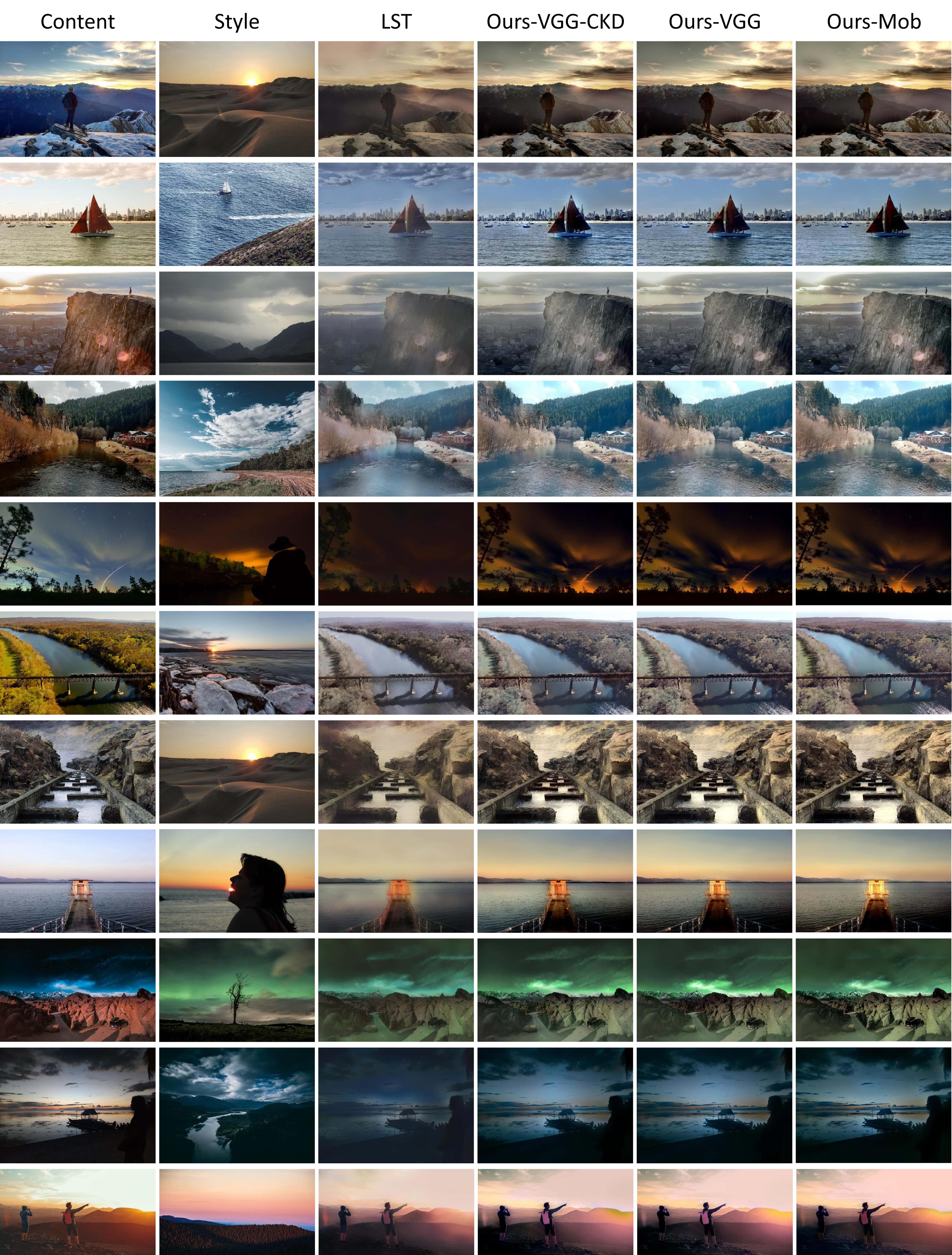}
    \caption{Comparison of the qualitative results from LST~\cite{li2019learning} and our PCA-distilled models. Compared to the results from our models, the results from LST are prone to blurred boundaries (low sharpness) and dullness (low contrast).}
    \label{fig:lst_qualitative}
\end{figure*}

First, we observe in \cref{fig:content_style_losses_all} that LST results in a content loss and a style loss both worse than those of CKD, which has the worst performance of all methods considered in the main paper. Qualitative results in \cref{fig:lst_qualitative} also show that compared to the results from our models, the results from LST are prone to blurred boundaries (low sharpness) and dullness (low contrast).

\newpage
\section*{Derivation of the equivalence of Equations 3 and 4 in the main paper}
In order to solve Equation 3 in the main paper with mini-batch gradient descent, we avoid the unstable minimization process due to the unbounded trace function by rewriting Equation 3 as Equation 4 where the objective is lower-bounded. Here we prove the equivalence of them (the following two equations).
\begin{equation}
    \max_{\mathbf{W}_{N,g}\mathbf{W}_{N,g}^\mathrm{T}=\mathbbm{1}} \frac{1}{M} \sum_{k=1}^M \mathrm{tr}(\mathbf{W}_{N,g}\mathbf{\bar{F}}_{N,k}\mathbf{\bar{F}}_{N,k}^\mathrm{T}\mathbf{W}_{N,g}^\mathrm{T}),
\end{equation}
and
\begin{equation}
    \min_{\mathbf{W}_{N,g}\mathbf{W}_{N,g}^\mathrm{T}=\mathbbm{1}}\frac{1}{M} \sum_{k=1}^M ||\mathbf{W}_{N,g}^\mathrm{T}\mathbf{W}_{N,g}\mathbf{\bar{F}}_{N,k} - \mathbf{\bar{F}}_{N,k}||^2_2.
\end{equation}

\begin{proof}
\begin{align}
\small
    &\min_{\mathbf{W}_{N,g}\mathbf{W}_{N,g}^\mathrm{T}=\mathbbm{1}}\frac{1}{M} \sum_{k=1}^M ||\mathbf{W}_{N,g}^\mathrm{T}\mathbf{W}_{N,g}\mathbf{\bar{F}}_{N,k} - \mathbf{\bar{F}}_{N,k}||^2_2 \\
    = & \min_{\mathbf{W}_{N,g}\mathbf{W}_{N,g}^\mathrm{T}=\mathbbm{1}}\frac{1}{M} \sum_{k=1}^M \text{tr}\big[
    \begin{aligned}
    &(\mathbf{W}_{N,g}^\mathrm{T}\mathbf{W}_{N,g}\mathbf{\bar{F}}_{N,k} - \mathbf{\bar{F}}_{N,k}) \\
    \cdot&(\mathbf{W}_{N,g}^\mathrm{T}\mathbf{W}_{N,g}\mathbf{\bar{F}}_{N,k} - \mathbf{\bar{F}}_{N,k})^\mathrm{T}
    \end{aligned}
    \big] \\
    = & \min_{\mathbf{W}_{N,g}\mathbf{W}_{N,g}^\mathrm{T}=\mathbbm{1}}\frac{1}{M} \sum_{k=1}^M \text{tr}\Bigg[
    \begin{aligned}
    &\mathbf{W}_{N,g}^\mathrm{T}\mathbf{W}_{N,g}\mathbf{\bar{F}}_{N,k}\mathbf{\bar{F}}_{N,k}^\mathrm{T}\mathbf{W}_{N,g}^\mathrm{T}\mathbf{W}_{N,g} \\
    - &\mathbf{W}_{N,g}^\mathrm{T}\mathbf{W}_{N,g}\mathbf{\bar{F}}_{N,k}\mathbf{\bar{F}}_{N,k}^\mathrm{T} \\
    - &\mathbf{\bar{F}}_{N,k}\mathbf{\bar{F}}_{N,k}^\mathrm{T}\mathbf{W}_{N,g}^\mathrm{T}\mathbf{W}_{N,g} \\
    +&\cancel{\mathbf{\bar{F}}_{N,k}\mathbf{\bar{F}}_{N,k}^\mathrm{T}}
    \end{aligned}
    \Bigg]. 
\end{align}
The last term $\mathbf{\bar{F}}_{N,k}\mathbf{\bar{F}}_{N,k}^\mathrm{T}$ is crossed out without affecting the optimization result. By using the identity that $\text{tr}[\mathbf{A}\mathbf{B}]$ $=$ $\text{tr}[\mathbf{B}\mathbf{A}]$ for any two multiplicable matrices $\mathbf{A}$ and $\mathbf{B}$, the objective can be further simplified as follows:
\begin{align}
    \small
    & \min_{\mathbf{W}_{N,g}\mathbf{W}_{N,g}^\mathrm{T}=\mathbbm{1}}\frac{1}{M} \sum_{k=1}^M \text{tr}\Bigg[
    \begin{aligned}
    &\mathbf{W}_{N,g}\mathbf{\bar{F}}_{N,k}\mathbf{\bar{F}}_{N,k}^\mathrm{T}\mathbf{W}_{N,g}^\mathrm{T}\mathbf{W}_{N,g}\mathbf{W}_{N,g}^\mathrm{T} \\
    - &\mathbf{W}_{N,g}\mathbf{\bar{F}}_{N,k}\mathbf{\bar{F}}_{N,k}^\mathrm{T}\mathbf{W}_{N,g}^\mathrm{T} \\
    - &\mathbf{W}_{N,g} \mathbf{\bar{F}}_{N,k}\mathbf{\bar{F}}_{N,k}^\mathrm{T}\mathbf{W}_{N,g}^\mathrm{T}
    \end{aligned}
    \Bigg]  \\
    = & \min_{\mathbf{W}_{N,g}\mathbf{W}_{N,g}^\mathrm{T}=\mathbbm{1}}\frac{1}{M} \sum_{k=1}^M \text{tr}\Bigg[
    \begin{aligned}
    &\mathbf{W}_{N,g}\mathbf{\bar{F}}_{N,k}\mathbf{\bar{F}}_{N,k}^\mathrm{T}\mathbf{W}_{N,g}^\mathrm{T}\mathbbm{1} \\
    - &2\mathbf{W}_{N,g}\mathbf{\bar{F}}_{N,k}\mathbf{\bar{F}}_{N,k}^\mathrm{T}\mathbf{W}_{N,g}^\mathrm{T} 
    \end{aligned}
    \Bigg] \\
    = & \min_{\mathbf{W}_{N,g}\mathbf{W}_{N,g}^\mathrm{T}=\mathbbm{1}}\frac{1}{M} \sum_{k=1}^M \text{tr}[- \mathbf{W}_{N,g}\mathbf{\bar{F}}_{N,k}\mathbf{\bar{F}}_{N,k}^\mathrm{T}\mathbf{W}_{N,g}^\mathrm{T} ] \\
    = & \max_{\mathbf{W}_{N,g}\mathbf{W}_{N,g}^\mathrm{T}=\mathbbm{1}}\frac{1}{M} \sum_{k=1}^M \text{tr}[ \mathbf{W}_{N,g}\mathbf{\bar{F}}_{N,k}\mathbf{\bar{F}}_{N,k}^\mathrm{T}\mathbf{W}_{N,g}^\mathrm{T} ].
\end{align}

\end{proof}

\section*{Derivation of the loss objective in Equation 6 in the main paper}
In the encoder distillation introduced in Section 3.2 in the main paper, we find using the feature reconstruction loss in \cref{eq:distillation_loss_supp} for encoder distillation results in a better convergence than directly taking $||\mathbf{\bar{F}}_{N,k}^e-\mathbf{W}_{N,g}\mathbf{\bar{F}}_{N,k}||^2_2$ as the loss function. 
\begin{equation}
    \mathcal{L}_{enc}^N(I_k) = ||\mathbf{W}_{N,g}^\mathrm{T}\mathbf{\bar{F}}_{N,k}^e - \mathbf{\bar{F}}_{N,k}||^2_2.
    \label{eq:distillation_loss_supp}
\end{equation}
Here we show the equivalence of \cref{eq:distillation_loss_supp} and $||\mathbf{\bar{F}}_{N,k}^e-\mathbf{W}_{N,g}\mathbf{\bar{F}}_{N,k}||^2_2$ as the loss function.

\begin{proof}
First,
\begin{align}
    &\min_{\mathbf{\bar{F}}_{N,k}^e}||\mathbf{\bar{F}}_{N,k}^e-\mathbf{W}_{N,g}\mathbf{\bar{F}}_{N,k}||^2_2 \\
    =& \min_{\mathbf{\bar{F}}_{N,k}^e} \text{tr}\big[(\mathbf{\bar{F}}_{N,k}^e-\mathbf{W}_{N,g}\mathbf{\bar{F}}_{N,k})(\mathbf{\bar{F}}_{N,k}^e-\mathbf{W}_{N,g}\mathbf{\bar{F}}_{N,k})^\mathrm{T}\big] \\
    =& \min_{\mathbf{\bar{F}}_{N,k}^e} \text{tr}\Bigg[
    \begin{aligned}
    &\mathbf{\bar{F}}_{N,k}^e(\mathbf{\bar{F}}_{N,k}^e)^\mathrm{T} + \cancel{ \mathbf{W}_{N,g}\mathbf{\bar{F}}_{N,k}\mathbf{\bar{F}}_{N,k}^\mathrm{T}\mathbf{W}_{N,g}^\mathrm{T}} \\ -&\mathbf{W}_{N,g}\mathbf{\bar{F}}_{N,k}(\mathbf{\bar{F}}_{N,k}^e)^\mathrm{T} - \mathbf{\bar{F}}_{N,k}^e\mathbf{\bar{F}}_{N,k}^\mathrm{T}\mathbf{W}_{N,g}^\mathrm{T}  
    \end{aligned} 
    \Bigg] \label{eq:direct_equiv},
\end{align}
where the term $\mathbf{W}_{N,g}\mathbf{\bar{F}}_{N,k}\mathbf{\bar{F}}_{N,k}^\mathrm{T}\mathbf{W}_{N,g}^\mathrm{T}$ can be crossed out since it does not contain the variable $\mathbf{\bar{F}}_{N,k}^e$ we optimize for.

Second,
\begin{align}
    &\min_{\mathbf{\bar{F}}_{N,k}^e}||\mathbf{W}_{N,g}^\mathrm{T}\mathbf{\bar{F}}_{N,k}^e - \mathbf{\bar{F}}_{N,k}||^2_2 \\
    = & \min_{\mathbf{\bar{F}}_{N,k}^e}\text{tr}\big[(\mathbf{W}_{N,g}^\mathrm{T}\mathbf{\bar{F}}_{N,k}^e - \mathbf{\bar{F}}_{N,k})(\mathbf{W}_{N,g}^\mathrm{T}\mathbf{\bar{F}}_{N,k}^e - \mathbf{\bar{F}}_{N,k})^\mathrm{T}\big] \\
    = & \min_{\mathbf{\bar{F}}_{N,k}^e}\text{tr}\Bigg[
    \begin{aligned}
    &\mathbf{W}_{N,g}^\mathrm{T}\mathbf{\bar{F}}_{N,k}^e(\mathbf{\bar{F}}_{N,k}^e)^\mathrm{T}\mathbf{W}_{N,g} + \cancel{\mathbf{\bar{F}}_{N,k}\mathbf{\bar{F}}_{N,k}^\mathrm{T}} \\
    -&\mathbf{W}_{N,g}^\mathrm{T}\mathbf{\bar{F}}_{N,k}^e\mathbf{\bar{F}}_{N,k}^\mathrm{T} - \mathbf{\bar{F}}_{N,k}(\mathbf{\bar{F}}_{N,k}^e)^\mathrm{T}\mathbf{W}_{N,g}
    \end{aligned}
    \Bigg]
\end{align}
By using the identity that $\text{tr}[\mathbf{A}\mathbf{B}]$ $=$ $\text{tr}[\mathbf{B}\mathbf{A}]$ for any two multiplicable matrices $\mathbf{A}$ and $\mathbf{B}$, the objective can be further simplified as follows:
\begin{equation}
    \min_{\mathbf{\bar{F}}_{N,k}^e}\text{tr}\Bigg[
    \begin{aligned}
    &\mathbf{\bar{F}}_{N,k}^e(\mathbf{\bar{F}}_{N,k}^e)^\mathrm{T}\cancel{\mathbf{W}_{N,g}\mathbf{W}_{N,g}^\mathrm{T}} \\
    -&\mathbf{\bar{F}}_{N,k}^e\mathbf{\bar{F}}_{N,k}^\mathrm{T}\mathbf{W}_{N,g}^\mathrm{T} - \mathbf{W}_{N,g}\mathbf{\bar{F}}_{N,k}(\mathbf{\bar{F}}_{N,k}^e)^\mathrm{T}
    \end{aligned}
    \Bigg],
    \label{eq:rewrite_equiv}
\end{equation}
where $\mathbf{W}_{N,g}\mathbf{W}_{N,g}^\mathrm{T}$ is crossed out since it is an identity matrix. Since the equality of \cref{eq:direct_equiv} and \cref{eq:rewrite_equiv}, we prove the equivalence of these two optimization problems $\min_{\mathbf{\bar{F}}_{N,k}^e}||\mathbf{\bar{F}}_{N,k}^e-\mathbf{W}_{N,g}\mathbf{\bar{F}}_{N,k}||^2_2$ and $\min_{\mathbf{\bar{F}}_{N,k}^e}||\mathbf{W}_{N,g}^\mathrm{T}\mathbf{\bar{F}}_{N,k}^e - \mathbf{\bar{F}}_{N,k}||^2_2$.
\end{proof}

\clearpage
\begin{figure*}[t]
    \centering
    \includegraphics[width=\textwidth]{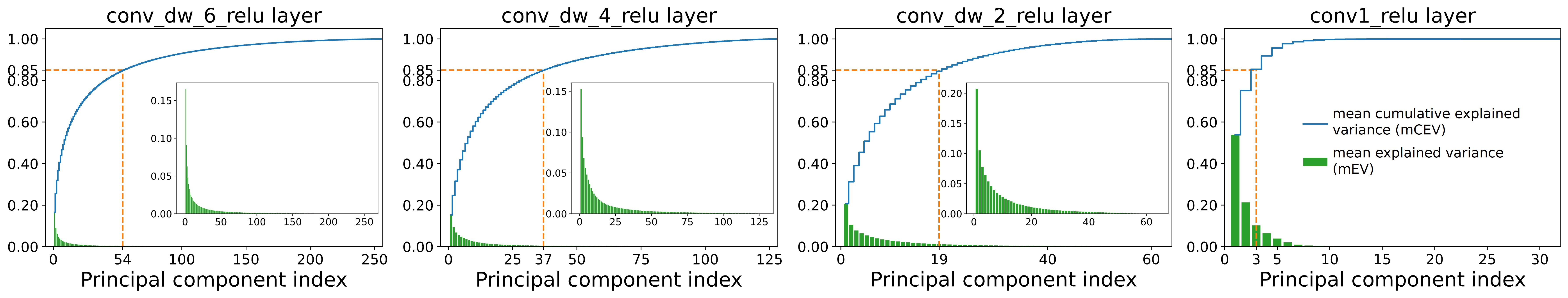}
    \caption{Mean explained variance (green histogram) and mean cumulative explained variance (blue curve) of the $\textit{conv\_dw\_relu}$ features and the $\textit{conv\_1\_relu}$ feature of MS-COCO images from MobileNet~\cite{howard2017mobilenets}. It is observed that on average $85\%$ of the variance of a $\textit{conv\_dw\_6\_relu}$, $\textit{conv\_dw\_4\_relu}$, $\textit{conv\_dw\_2\_relu}$, or $\textit{conv\_1\_relu}$ feature can be explained by $54$, $37$, $19$, or $3$ eigenvectors of the feauture covariance, respectively.}
    \label{fig:explained_variance_mob}
\end{figure*}
\section*{Channel length selection for Ours-Mob model distilled from MobileNet}
To demonstrate the generalizability of our PCA knowledge distillation, we apply it to distill style information from MobileNet~\cite{howard2017mobilenets}. By following how we select the layers from VGG-19~\cite{simonyan2014very} for style representation: we select the layer right after each downsampling layer for four layers in total, we select the $\textit{conv\_dw\_6\_relu}$, $\textit{conv\_dw\_4\_relu}$, $\textit{conv\_dw\_2\_relu}$, and $\textit{conv\_1\_relu}$ layers from MobileNet for style representation.

We follow the same procedure as described in Section 4.1 in the main paper to distill style information from the selected layers to a smaller model which we call Ours-Mob in the main paper. We plot mCEV and mEV for each selected layer in \cref{fig:explained_variance_mob}. It is observed that on average $85\%$ of the variance of a $\textit{conv\_dw\_6\_relu}$, $\textit{conv\_dw\_4\_relu}$, $\textit{conv\_dw\_2\_relu}$, or $\textit{conv\_1\_relu}$ feature can be explained by $54$, $37$, $19$, or $3$ eigenvectors of the feauture covariance, respectively. However, we find that if $C_1^e$ is set to $3$, it prevents the distillation from the $\textit{conv\_dw\_2\_relu}$ layer if $C_2^e$ is set to $19$. We fix this issue by following the value of $C_1^e$ we use in Ours-VGG. In the end, we set four channel lengths ($C_1^e$, $C_2^e$, $C_3^e$, $C_4^e$) to be (10, 19, 37, 54), resulting in Ours-Mob model.

\section*{High-frequency residuals improve the high-frequency detail construction of PhotoWCT and the original CKD-distilled model}
\begin{figure*}[t]
    \centering
    \includegraphics[width=\textwidth]{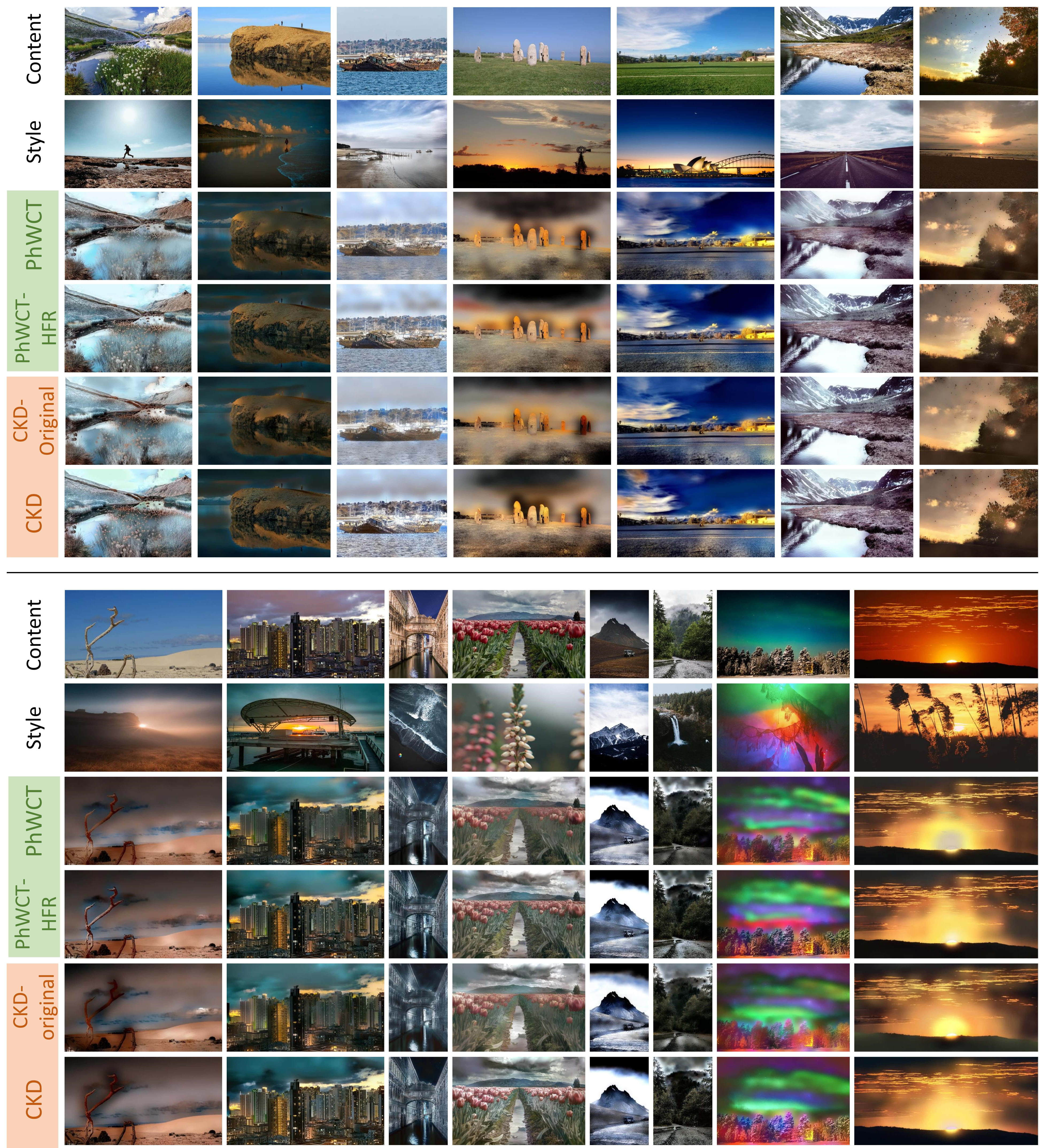}
    \caption{High-frequency residuals (HFR)~\cite{chiu2021photowct2} improve the content preservation of PhotoWCT~\cite{li2018closed} and CKD-original~\cite{wang2020collaborative} by reinforcing the high-frequency detail construction.}
    \label{fig:qualitative_hfr}
\end{figure*}

Recall in the Section 2 in the paper that we mentioned PhotoWCT~\cite{li2018closed} is poor at preserving content due to two reasons: too strong stylization strength that introduces artifacts and the lossy architecture that does not hold the high-frequency detail well. To consider stylization strength as the main factor that affect the content preservation, we fix the lossy architecture by introducing high-frequency residuals~\cite{chiu2021photowct2} (HFR) to PhotoWCT. The resulting model which we call PhotoWCT-HFR reduces the content loss of PhotoWCT by $17.7\%$ (\cref{fig:content_style_losses_all}). The better content preservation of PhotoWCT-HFR is due to its better high-frequency construction as exemplified in \cref{fig:qualitative_hfr}.

Similarly, the original model distilled with CKD~\cite{wang2020collaborative} (which we call CKD-original) is for artistic style transfer and also poor at constructing high-frequency details. To have a fair comparison, we again introduce HFR to our CKD-distilled model.  The resulting model which we call CKD in the main paper reduces the content loss of CKD-original by $25.2\%$ (\cref{fig:content_style_losses_all}). The better content preservation of CKD is due to its better high-frequency construction as exemplified in \cref{fig:qualitative_hfr}.

\section*{Demonstration that CKD does not improve performance when implemented using our PCA-derived channel lengths instead of the empirical ones}

\begin{figure*}[t]
    \centering
    \includegraphics[width=\textwidth]{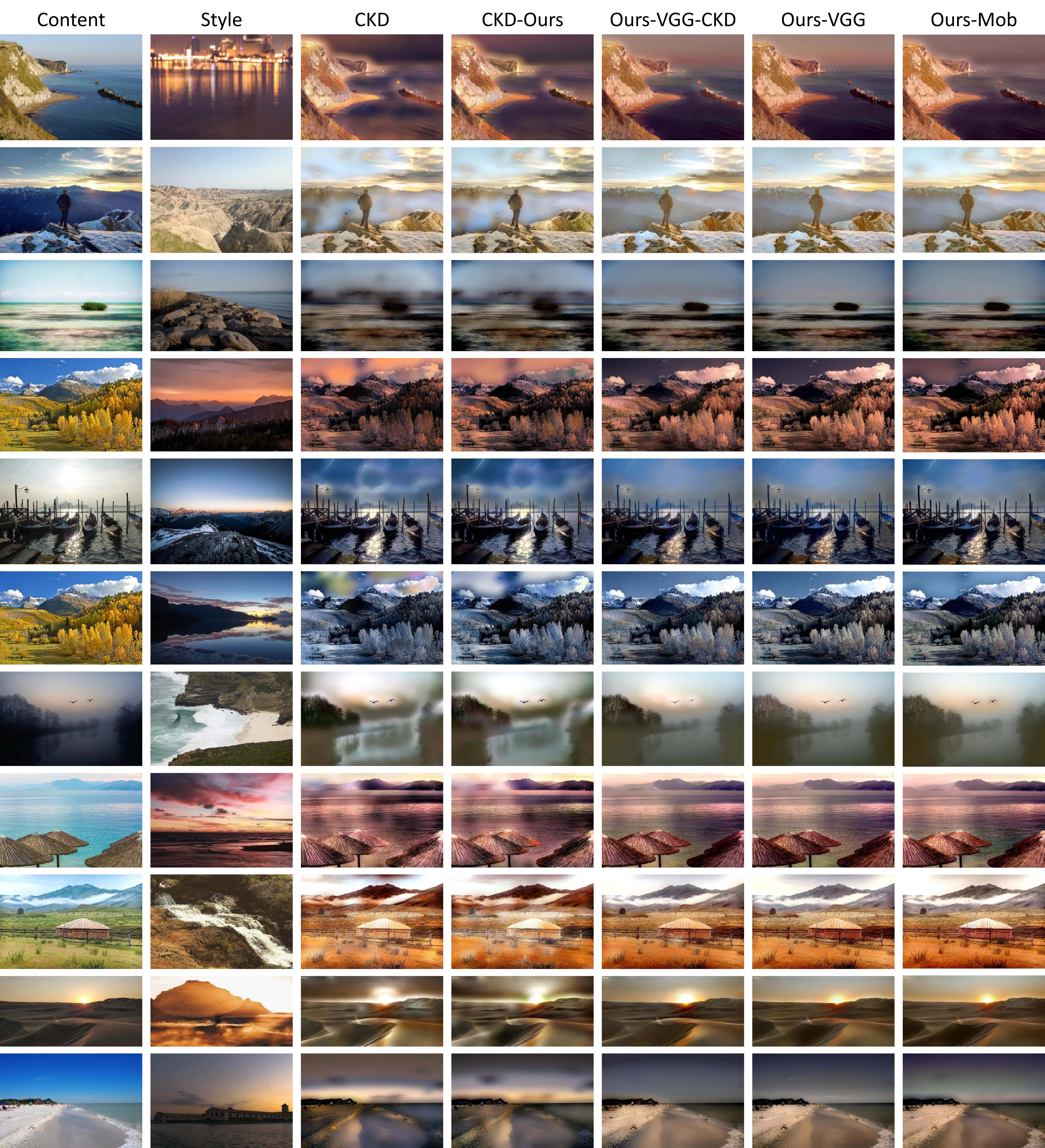}
    \caption{Qualitative comparison between our models, CKD and CKD-Ours. We observe that CKD and CKD-Ours produce very similar results with lots of artifacts, and our models consistently result in more photorealistic images than both CKD-distlled models.}
    \label{fig:qualitative_ckd}
\end{figure*}

Unlike our PCA distillation, which has clear guidelines for channel length selection, the previous method CKD~\cite{wang2020collaborative} empirically sets the channel lengths to be ($C_1^e=16$, $C_2^e=32$, $C_3^e=64$, $C_4^e=128$) when distilling from VGG-19 and results in the CKD model. We show here that the model distilled with CKD using our smaller channel lengths ($C_1^e=10$, $C_2^e=20$, $C_3^e=58$, $C_4^e=64$), which we call CKD-Ours, does not change the performance of CKD as shown in \cref{fig:content_style_losses_all}. Qualitatively, as shown in \cref{fig:qualitative_ckd}, we observe that both CKD and CKD-Ours produce lots of artifacts in the synthesized images, and our models consistently result in more photorealistic images than both CKD-distlled models.

\clearpage

\clearpage

\clearpage
\section*{Inference time of different models on the CPU}
we report in \cref{tab:model_charateristics_supp} models' inference times on an Intel Xeon W-2195 CPU @ 2.30GHz with workstation memory of 256GB. The results have a similar trend to those in Table 1 in the main paper: our distilled models achieve the fastest inference time. Moreover, while the non-distilled models spend 1-2+ minutes rendering an 8K image with CPUs, our models require considerably less time; i.e., 10 seconds/image. 

\setlength{\tabcolsep}{1.8pt}
\begin{table}[h!]
    \centering\small
    \def\arraystretch{1.2}
    \begin{tabular}{l c c c c c c}
    \toprule
        Model & HD & FHD & QHD & 4K & 5K & 8K \\
    \midrule
    WCT$^2$ & 5.92 & 18.04 & 31.61 & 71.28 & 109.13 & X\\
    PhotoWCT-HFR & 4.91 & 10.31 & 18.06 & 38.44 & 69.22 & 149.22\\
    PhotoWCT$^2$ & 2.33 & 5.06 & 9.08 & 20.39 & 36.15 & 79.74 \\
    \rowcolor{lightgray!40}[\dimexpr\tabcolsep+0.1pt\relax] CKD & 1.04 & 2.20 & 3.80 & 8.03 & 14.66 & 34.16\\
    \rowcolor{lightgray!40}[\dimexpr\tabcolsep+0.1pt\relax] Ours-VGG-CKD & 0.47 & 0.96 & 1.68 & 3.60 & 6.51 & 14.21 \\\noalign{\vskip-0.25pt}
    \rowcolor{lightgray!40}[\dimexpr\tabcolsep+0.1pt\relax] Ours-VGG & \bf{0.40} & \bf{0.77} & \bf{1.44} & \bf{2.88} & \bf{4.84} & \bf{10.78} \\\noalign{\vskip-0.25pt}
    \hdashline
    \rowcolor{lightgray!40}[\dimexpr\tabcolsep+0.1pt\relax] Ours-Mob & \bf{0.21} & \bf{0.37} & \bf{0.76} & \bf{1.62} & \bf{2.76} & \bf{6.02} \\\noalign{\vskip-0.25pt}
    \bottomrule
    \end{tabular}
    \vspace{-0.5em}
    \caption{Inference time (s/img) of different models on CPU. The naming follows the main paper. X: segmentation fault (We found this is not due to the common reason of insufficient memory or system stack size. The reason remains unknown).}
    \label{tab:model_charateristics_supp}
\end{table}

\section*{More stylized images from the PST dataset}
We show several qualitative results from the PST dataset~\cite{xia2020joint} in Figure 5 in the main paper. Here we show more of them in \cref{fig:qualitative_pst}.

\section*{Global eigenbases reflect style better than local eigenbases}
\begin{figure*}
    \centering
    \includegraphics[width=0.8\textwidth]{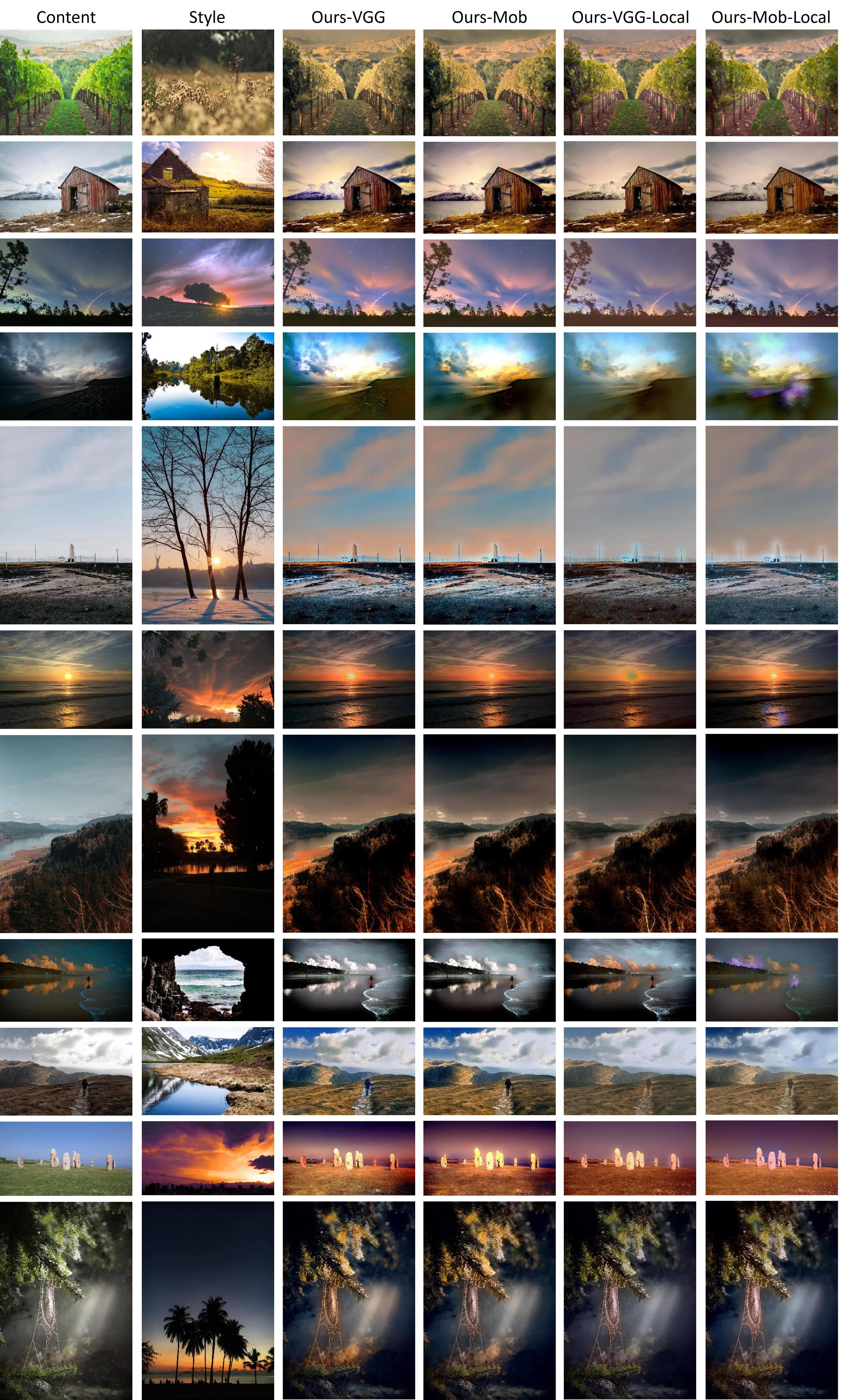}
    \caption{PCA distillation with global eigenbases reflects better style than local eigenbases.}
    \label{fig:local_vs_global}
\end{figure*}

\begin{figure*}[t]
    \centering
    \includegraphics[width=\textwidth]{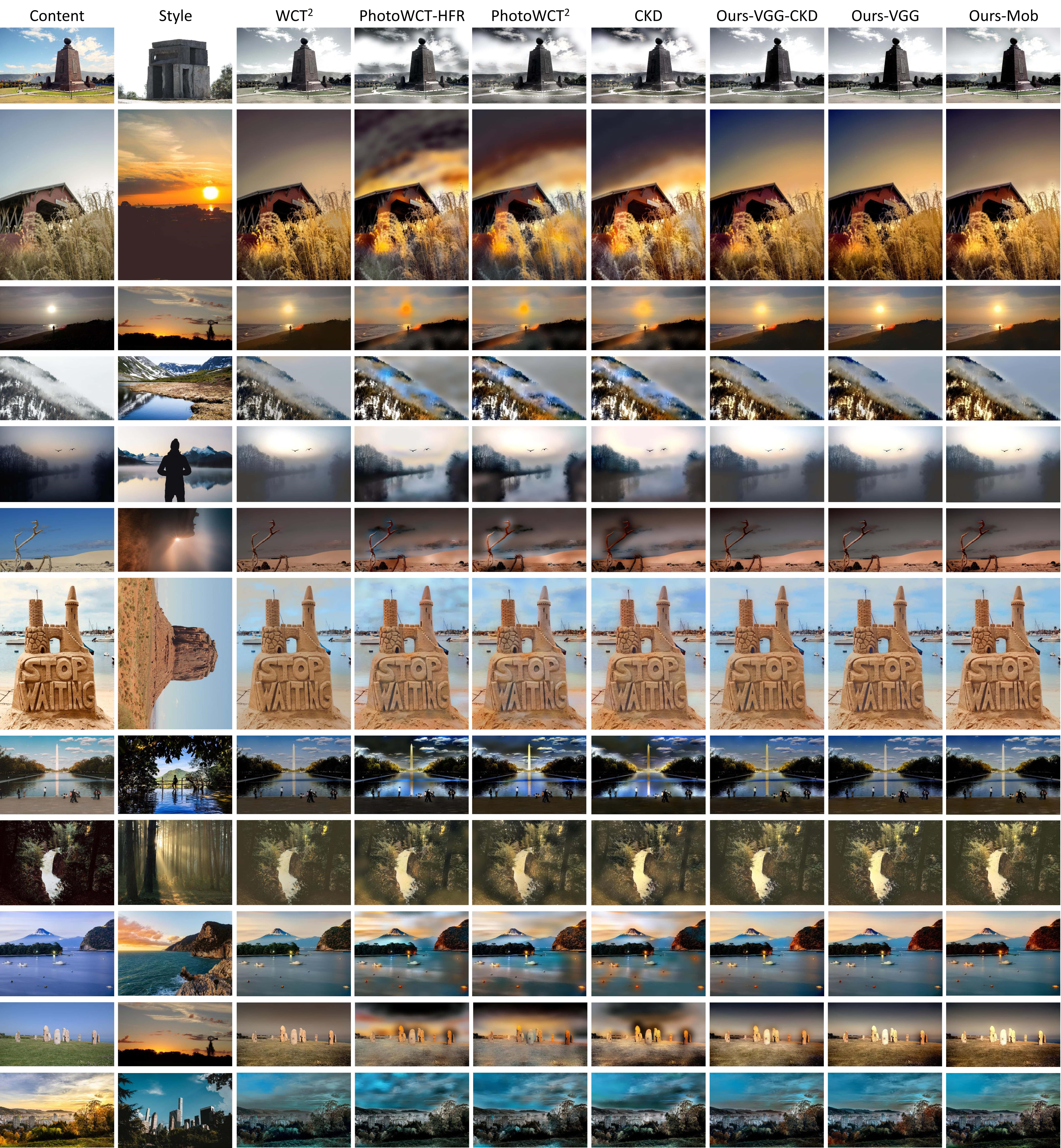}
    \caption{More stylized images from the PST dataset~\cite{xia2020joint}. This figure expands the Figure 5 in the main paper.}
    \label{fig:qualitative_pst}
\end{figure*}

In Section 3 of the main paper, we explain the limitation of local eigenbases to faithfully reflect the style of style images. To overcome this, we propose using global eigenbases. The quantitative result (\cref{fig:content_style_losses_all}) justifies this limitation and our strategy. We show in \cref{fig:local_vs_global} some qualitative results comparing the impact of global and local eigenbases.

\section*{Results of stylized images for 4K+ resolutions from our PCA-distilled models}

\begin{figure*}[t]
    \centering
    \includegraphics[width=\textwidth]{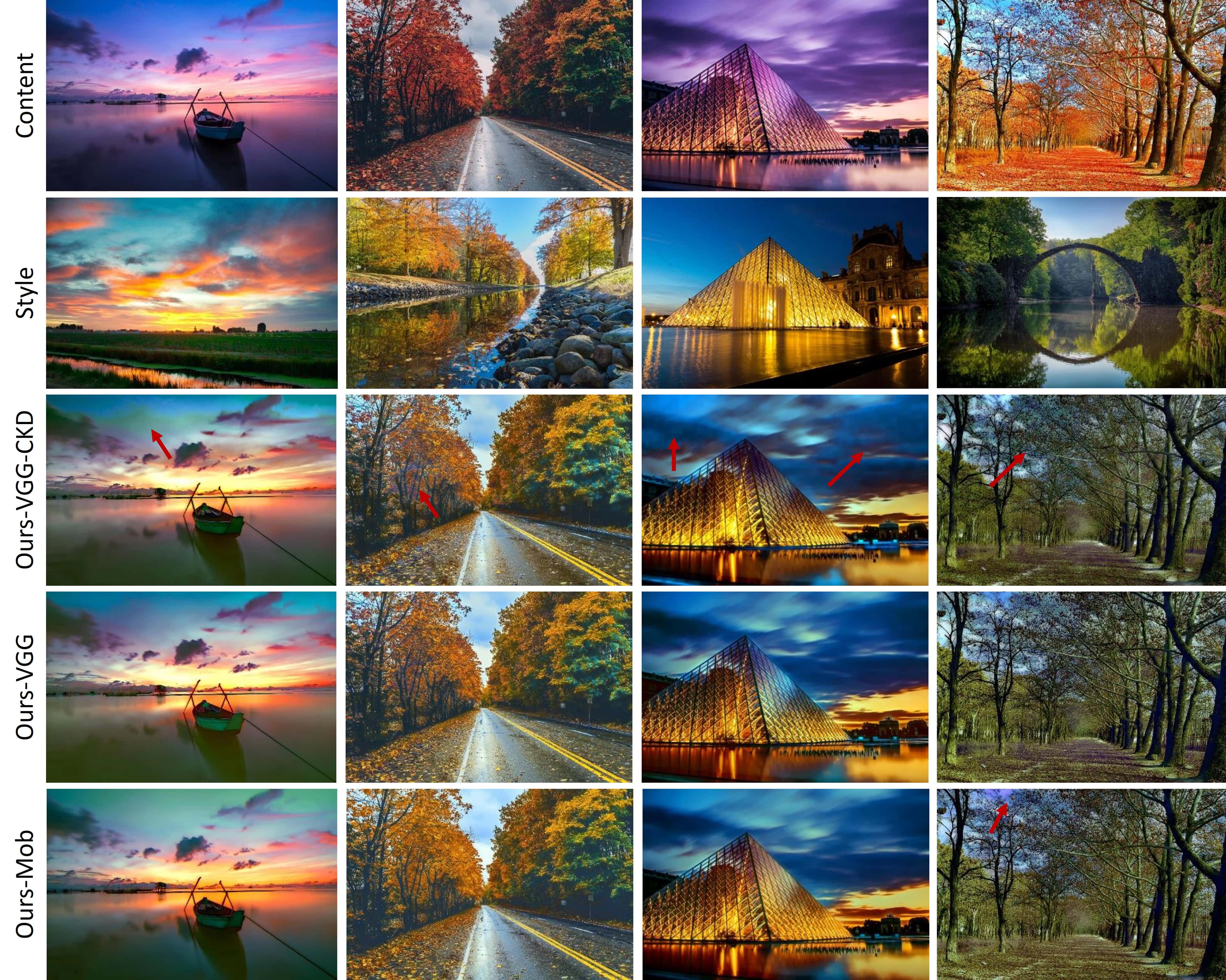}
    \caption{Stylized images of 4K+ resolutions resulting from our models. Several artifacts are pointed out with the red arrows. (Part 1/4)}
    \label{fig:4k_1}
\end{figure*}

\begin{figure*}[t]
    \centering
    \includegraphics[width=\textwidth]{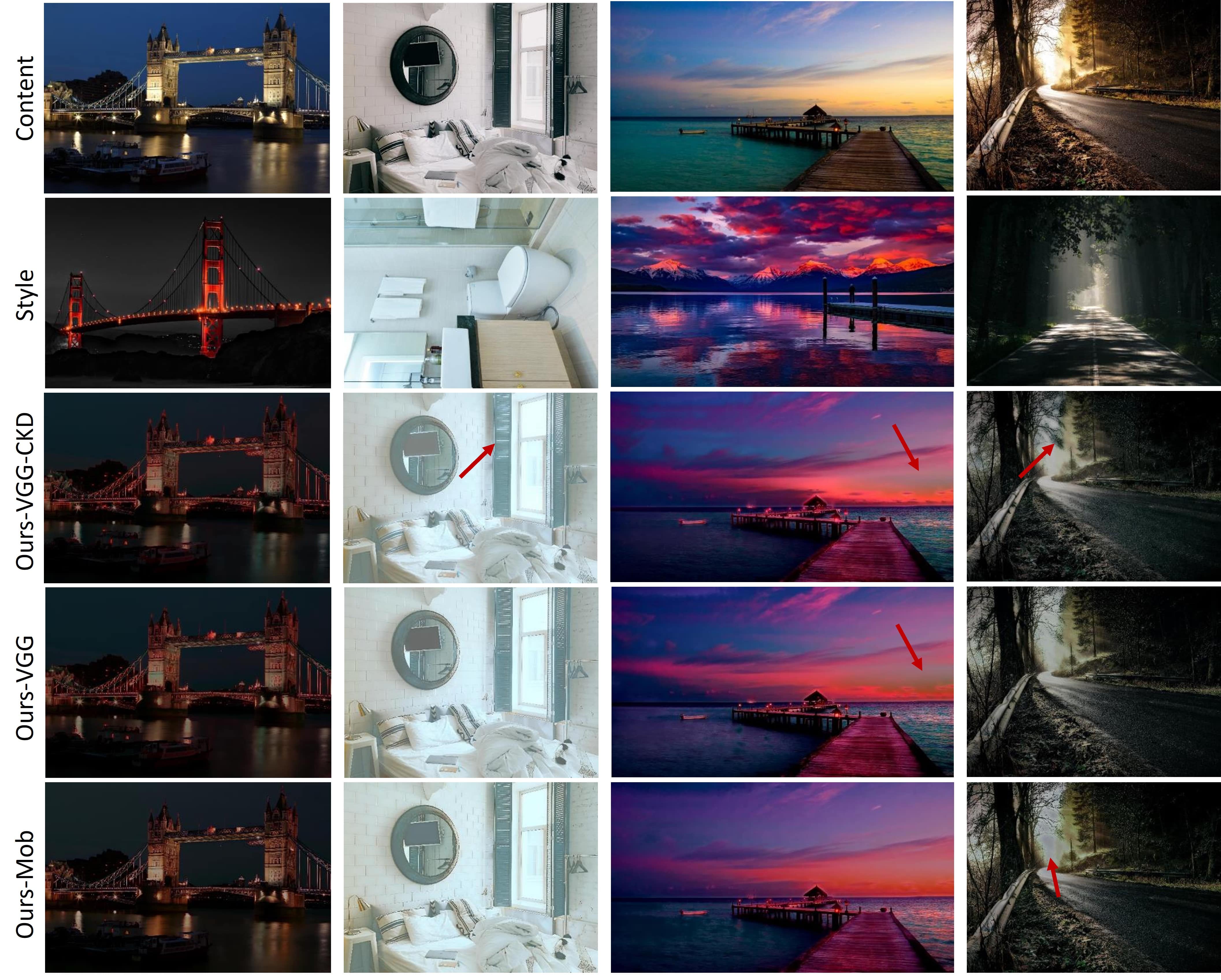}
    \caption{Stylized images of 4K+ resolutions resulting from our models. Several artifacts are pointed out with the red arrows. (Part 2/4)}
    \label{fig:4k_2}
\end{figure*}

\begin{figure*}[t]
    \centering
    \includegraphics[width=\textwidth]{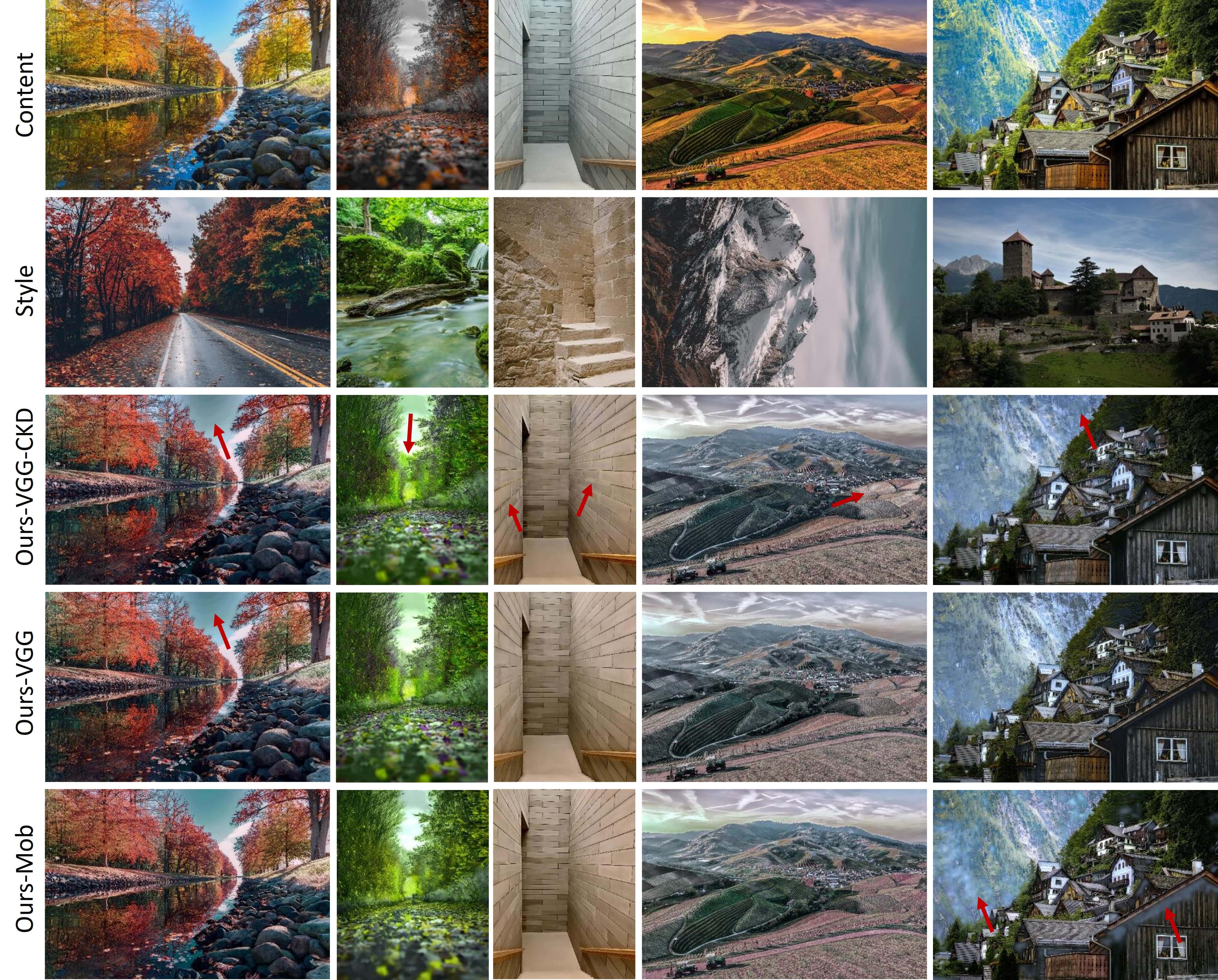}
    \caption{Stylized images of 4K+ resolutions resulting from our models. Several artifacts are pointed out with the red arrows. (Part 3/4)}
    \label{fig:4k_3}
\end{figure*}

\begin{figure*}[t]
    \centering
    \includegraphics[width=\textwidth]{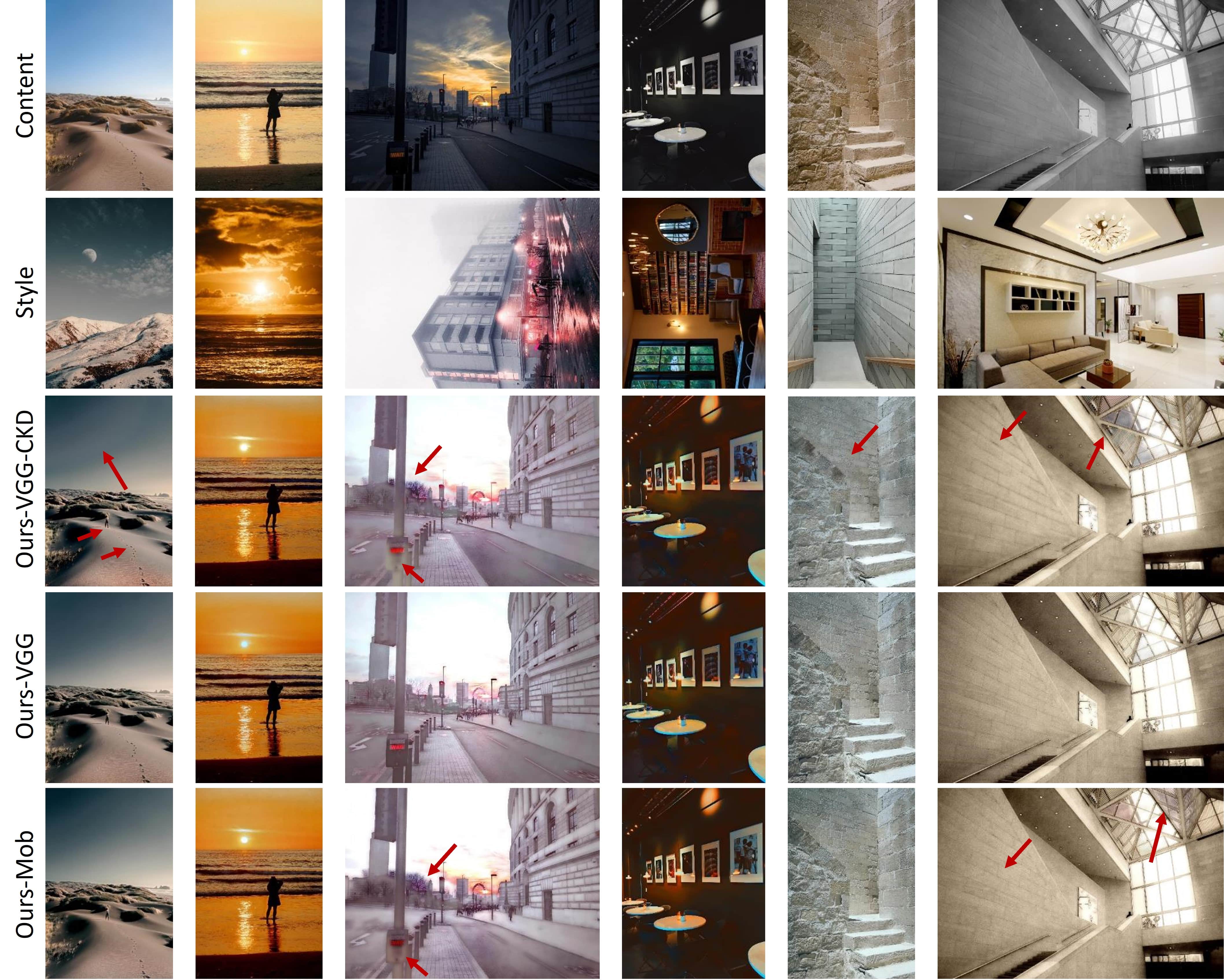}
    \caption{Stylized images of 4K+ resolutions resulting from our models. Several artifacts are pointed out with the red arrows.  (Part 4/4)}
    \label{fig:4k_4}
\end{figure*}

Since images of large resolutions (e.g. 4K and beyond) contain more high-frequency details, which are hotbeds for artifacts to form in stylization, the ability to reduce artifacts of a photorealistic style transfer model can be better manifested in the stylization of images of 4K+ resolutions. We show such results from our PCA-distilled models in \cref{fig:4k_1}, \cref{fig:4k_2}, \cref{fig:4k_3}, and \cref{fig:4k_4}. We notice that Ours-VGG, which uses the channel lengths derived with PCA, constantly preserve better content than Ours-VGG-CKD, which uses the channel lengths empirically selected in the CKD paper~\cite{wang2020collaborative}, by trimming off the slight artifacts. We also notice that while Ours-Mob results in a slightly lower content loss than Ours-VGG in the stylization of images of lower resolutions (images in PST dataset~\cite{xia2020joint}) as shown in \cref{fig:content_style_losses_all}, Ours-VGG produces fewer artifacts than Ours-Mob in the stylization of our 4K+ images. 

\clearpage

\clearpage

\clearpage

\clearpage

\clearpage

\clearpage

\newpage

{\small
\bibliographystyle{ieee_fullname}
\bibliography{egbib}
}

\end{document}